\documentclass[runningheads]{llncs}
\usepackage[T1]{fontenc}
\usepackage{graphicx}
\usepackage{float}

\usepackage{amsmath}
\usepackage{amssymb}
\usepackage{booktabs}
\usepackage{algorithm, algpseudocode}
\usepackage{url}
\usepackage{caption}
\usepackage{subcaption}
\usepackage{xcolor}
\usepackage{paralist}
\usepackage{cite}
\usepackage{hyperref}

\makeatletter
\def\blfootnote#1{{\xdef\@thefnmark{}\@footnotetext{#1}}}
\makeatother

\begin{document}
\title{Multi-Task Optimization over Networks of Tasks}

%
%
\author{Julian Hatzky\inst{1}  \and
Thomas Bartz-Beielstein\inst{2} \and
A. E. Eiben\inst{1}  \and
Anil Yaman\inst{1}}
\authorrunning{J. Hatzky et al.}
%
\institute{Vrije Universiteit Amsterdam, Amsterdam, The Netherlands\\
\email{\{j.hatzky, a.e.eiben, a.yaman\}@vu.nl} \and
TH Köln, Cologne, Germany\\
\email{thomas.bartz-beielstein@th-koeln.de}}
\maketitle              
\blfootnote{\scriptsize Accepted at PPSN 2026.}

\begin{abstract}



Multi-task optimization is a powerful approach for solving a large number of tasks in parallel.
However, existing algorithms face distinct limitations:
Population-based methods scale poorly and remain underexplored for large task sets.
Approaches that do scale beyond a thousand tasks are mostly MAP-Elites variants and rely on a fixed, discretized archive that disregards the topology of the task space.
We introduce MONET (Multi-Task Optimization over Networks of Tasks), a multi-task optimization algorithm that models the task space as a graph: tasks are nodes, and edges connect tasks in the task parameter space. This representation enables knowledge transfer between tasks and remains tractable for high-dimensional problems while exploiting the topology of the task space. MONET combines \emph{social learning}, which generates candidates from neighboring nodes via crossover, with \emph{individual learning}, which refines a node's own solution independently via mutation.
We evaluate MONET on four domains (archery, arm, and cartpole with
5,000 tasks each; hexapod with 2,000 tasks) and show that it matches or
exceeds the performance of existing MAP–Elites-based baselines across
all four domains.

\end{abstract}


\section{Introduction}\label{lab:intro}

To maintain reliable performance, real-world embodied agents must remain operational even when their configurations change. For example, in scenarios where a legged robot loses a limb~\cite{MAP_Elites_NaturePaper,bongard2006resilient} or a modular robot reconfigures itself to navigate a narrow passage~\cite{yim2007modular,vandiggelen2024modelfree,samarakoon2022metaheuristic}, they are required to maintain their performance. In such cases, the ability to adapt an existing control strategy instead of relearning a new strategy from scratch can prove to be more efficient. It is possible to achieve this by employing the multi-task optimization approach~\cite{gupta2015multifactorial}, where each configuration variant (e.g. morphology) is treated as a distinct task, and solutions (e.g. controller) are pre-computed for all of them, so that when the configurations change, an appropriate controller is already available or is faster to adapt~\cite{DBLP:journals/corr/MouretC15,chatzilygeroudis2021qd}. In other words, given a parameterized family of tasks, the multi-task optimization approach aims to find a high-quality solution for each task simultaneously and in parallel~\cite{gupta2015multifactorial}.


\begin{figure}
  \centering
  \includegraphics[width=\textwidth]{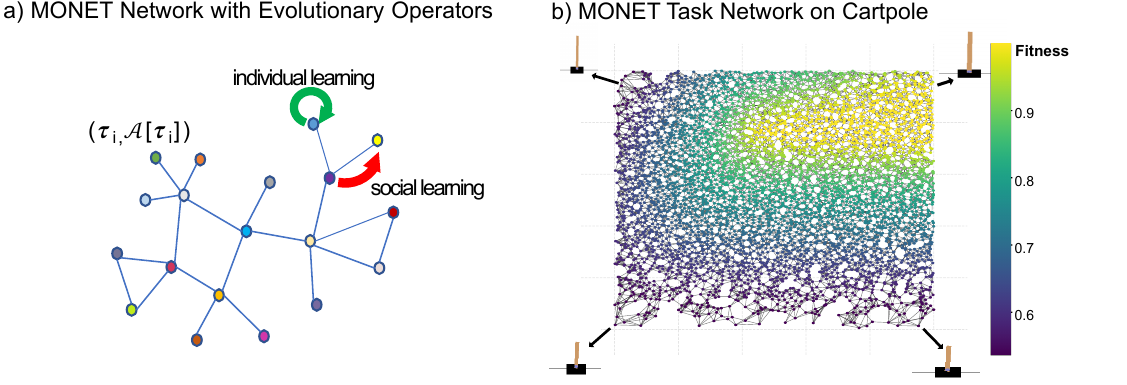}
  \caption{Illustration of the MONET task network structure. Each node $V_i = \boldsymbol{\tau}_i$ represents a task, labelled by its current best solution $\mathcal{A}[\boldsymbol{\tau}_i] \in \Theta$. (a) Task network with the individual and social learning strategies that operate on it. (b) Task network and solution qualities at $500{,}000$ evaluations ($50\%$ of the total).}
  \label{fig:teaser}
\end{figure}

Existing multi-task optimization algorithms face a fundamental scalability--structure tradeoff. Population-based multi-task evolutionary methods~\cite{huynhthithanhEnsembleMultifactorialEvolutionBiasedSkillFactorInheritanceManyTaskOptimization2023,wangSolvingMultitaskOptimizationProblemsAdaptiveKnowledgeTransferAnomalyDetection2022} maintain a separate population per task and quickly become impractical as the number of tasks grows into the thousands and beyond. Quality-diversity (QD) algorithms and MAP-Elites~\cite{MAP_Elites_NaturePaper,DBLP:journals/corr/MouretC15} in particular, bypass this by replacing per-task populations with a single structured archive. MAP-Elites has been extended to multi-task settings by Multi-Task MAP-Elites (MT-ME)~\cite{mouretQualitydiversitymultitaskoptimization2020}, which solves a fixed set of tasks via a globally shared archive, and Parametric-Task MAP-Elites (PT-ME)~\cite{anneParametricTaskMAPElites2024}, which extends this to continuous task spaces. Both have demonstrated strong performance on problems with thousands of tasks.

However, these archive-based methods exhibit two intertwined limitations. First, the archive is a \emph{fixed discretized grid} over the task or feature space: with $c$ bins per dimension and a $d$-dimensional task space its size grows as $\mathcal{O}(c^d)$, so the representation becomes infeasible already for moderate $d$. Second, the archive is largely \emph{topologically flat}: in MT-ME any solution can serve as a parent for any task. PT-ME partially addresses this through a local linear regression operator that exploits Delaunay triangulation~\cite{delaunay1934} between tasks to calculate an adjacency matrix. Even so, this locality is confined to one of two variation paths, and the global Simulated Binary Crossover (SBX) operator~\cite{SBX} still draws parents without regard for task similarity. Neither method makes the task-space topology a first-class structural element that governs all recombination.

We introduce MONET (Multi-Task Optimization over Networks of Tasks), a multi-task optimization algorithm that resolves both limitations by replacing the flat archive with an explicit graph over tasks (Figure~\ref{fig:teaser}): each node in the vertex set $\mathcal{V}$ represents a task and stores its current best solution, while bidirectional edges in $\mathcal{E}$ connect tasks to enable knowledge transfer (e.g., through crossover or direct copy). Unlike a grid, this representation does not depend on a discretization of the task space and scales as $\mathcal{O}(|\mathcal{V}|\,k)$, where $k$ is the number of neighbors per node, remaining tractable as the task space dimensionality grows. In contrast to a flat archive, the graph encodes topology explicitly: every node has a well-defined neighborhood that the search can exploit. The evolutionary process operates on this graph through two complementary mechanisms. \emph{Individual learning} improves a solution in isolation through mutation, exploring the local fitness landscape of a single task. \emph{Social learning} transfers information between neighboring nodes in variable ratios by selecting parent solutions from neighboring tasks and generating offspring through crossover. This separation is inspired by cultural evolution, where individual innovation and the social spread of useful innovations through a population jointly drive adaptation~\cite{rendell2010socialLearning,yaman2021distributed}.

The balance between individual and social learning, the strategy to select a node (random or best fit), the size of the neighborhood, and the type of neighborhood (closest neighbors, distance-proportional sampling, or random selection) are key design choices that shape how information flows through the graph. A small neighborhood with strong social learning concentrates exploitation among very similar tasks, while a larger neighborhood or weaker social learning permits broader exploration. Understanding which of these choices matters and whether a single robust configuration exists in different problem domains is essential for the practical application of MONET.

To this end, we conducted a hyperparameter study in four environments: archery, arm, and cartpole with 5,000 tasks each, and hexapod with 2,000 tasks. We show that MONET matches or outperforms both MT-ME and PT-ME across all four environments, demonstrating that exploiting task-space structure through a network of task neighborhoods is an effective approach to multi-task optimization at scale.
Our contributions are:
\begin{compactenum}
    \item A unified formal taxonomy of MAP-Elites variants for multi-task optimization that exposes the structural modifications distinguishing each variant and motivates MONET (Section~\ref{lab:related}).
    \item MONET, a graph-based multi-task optimizer that makes task-space neighborhoods an explicit element of recombination and combines localized social and individual learning (Section~\ref{lab:methods}).
    \item A systematic hyperparameter study across four task domains (archery, arm, cartpole, and hexapod) analyzing the effects of neighborhood type, neighborhood size, and the social/individual balance (Sections~\ref{lab:experimental} and~\ref{lab:results}).
    \item Empirical evidence that a single MONET configuration generalizes across all four domains and matches or outperforms MT-ME and PT-ME, showing task-space topology to be a valuable and underexploited source of information in multi-task optimization (Sections~\ref{lab:results} and~\ref{lab:discussion}).
    
\end{compactenum}

\section{Related Work}\label{lab:related}

Evolutionary multi-task optimization has become a highly active area of research within the computational intelligence community \cite{gupta2015multifactorial,EMT_6_applications,multi_task_survey,osaba2021evolutionarymultitaskoptimizationmethodological}.
To concretize the description, we specify the problem setting as follows.
Consider a set of $n$ tasks, each described by a parameter vector $\boldsymbol{\tau}_i \in \mathcal{T}$, for which we want to find the corresponding solutions $\boldsymbol{\theta}_1, \ldots, \boldsymbol{\theta}_n \in \Theta$ simultaneously. The goal of multi-task optimization is to find, for every task $\boldsymbol{\tau} \in \mathcal{T}$, an optimal solution as:
\begin{equation}
\boldsymbol{\theta}^{*}_{\boldsymbol{\tau}} = \arg\max_{\boldsymbol{\theta} \in \Theta} f(\boldsymbol{\theta}, \boldsymbol{\tau}),
\label{eq:multitask}
\end{equation}
where $\Theta$ is the solution space, $\mathcal{T}$ the task parameter space and $f: \Theta \times \mathcal{T} \rightarrow \mathbb{R}$ the fitness function.

Existing methodologies generally fall into two categories: \emph{implicit transfer}, which utilizes a single, unified search space for all tasks, and \emph{explicit transfer}, which operates across distinct solution spaces. Although implicit transfer is relatively straightforward to implement, its application is largely restricted to homogeneous environments. In contrast, explicit transfer accommodates heterogeneous tasks, though determining the optimal knowledge to exchange between distinct spaces remains a complex challenge. To facilitate this transfer, contemporary frameworks have leveraged probabilistic search distributions, search direction vectors, higher-order heuristics, and surrogate models. A representative example is the Multi-Factorial Evolutionary Algorithm II (MFEA-II)~\cite{bali2020mfea2}, which learns an online inter-task transfer parameter to modulate the probabilistic crossover between tasks and thereby suppress negative transfer during multifactorial search.

Historically, the application of these methods has been limited to optimizing a small number of tasks concurrently (typically between two and ten). As catalogued in the evolutionary multi-tasking (EMT) surveys of~\cite{EMT_6_applications, osaba2021evolutionarymultitaskoptimizationmethodological}, examples include optimizing three robot morphologies via \emph{neuroevolution}~\cite{bali2020mfea2}, two symbolic regression datasets using genetic programming~\cite{EMT_symbolic_regression_with_GP}, ten numerical functions for software testing~\cite{EMT_software_testing_multifactorial}, multi-UAV path planning across three tasks~\cite{EMT_UAV_path_planning}, power dispatch optimization for two bus systems~\cite{EMT_complex_design_electric_power_dispatch}, logistics planning for four package delivery problems~\cite{EMT_logistics_planning_combinatorial_optimzation}, and the design of three diode models~\cite{EMT_complex_design_photovoltaic_models}. Similar multi-task formulations exist in other fields, such as Bayesian optimization~\cite{Multi_task_bayesian_optimization}, which is frequently used to tune machine learning models across multiple datasets. However, because Bayesian approaches rely on Gaussian processes that scale cubically\footnote{Methods like the Nyström approximation alleviate the cubic scaling, though not without trade-offs.}, they are restricted to computationally expensive fitness functions and struggle to scale beyond a thousand evaluations~\cite{Bayesian-Optimization_review}.
To address these scalability limits, recent works in evolutionary many-task optimization have successfully tackled increasingly larger problem sets, scaling to 30, 50, 500, and up to 2,000 tasks simultaneously. MONET aligns with this trajectory toward large-scale optimization.

In order to show the connection of the proposed method to the existing literature, we first introduce a unified notation framework for the whole MAP-Elites algorithm family, enabling each multi-task variant to be presented as a modification of its underlying formulation\footnote{We adopt the maximization convention throughout; a minimization problem with cost $c$ is handled by setting $f = -c$.}. The same base is then extended once more in Section~\ref{lab:methods} when MONET is defined.

\begin{definition}[MAP-Elites]\label{def:mapelites}
Let $\Theta$ denote a solution space, $f : \Theta \to \mathbb{R}$ a fitness function to be \emph{maximized}, $\mathcal{B}$ a behavior space, and $\mathbf{b} : \Theta \to \mathcal{B}$ a behavioral descriptor that maps a solution to its behavior. Fix a finite discretization of $\mathcal{B}$ into cells. MAP-Elites~\cite{MAP_Elites_NaturePaper,DBLP:journals/corr/MouretC15} maintains an archive $\mathcal{A}$\footnote{We use uppercase calligraphic $\mathcal{A}$ because the archive is primarily a data structure that is \emph{treated as} a partial function; this convention is standard in the MAP-Elites literature and consistent with the use of uppercase for structured objects such as operators or transforms.}, treated as a partial function from cells to $\Theta$, initialized empty and updated by the following elitist rule: given a candidate $\boldsymbol{\theta} \in \Theta$, insert $\boldsymbol{\theta}$ into the cell containing $\mathbf{b}(\boldsymbol{\theta})$ iff that cell is empty or $f(\boldsymbol{\theta}) > f(\mathcal{A}[\mathbf{b}(\boldsymbol{\theta})])$. The run returns $\mathcal{A}$, which trades objective value against behavioral coverage rather than optimizing $f$ alone.
\end{definition}

\begin{figure}[t]
  \centering
  \begin{subfigure}[t]{0.48\textwidth}
    \centering
    \includegraphics[width=\textwidth]{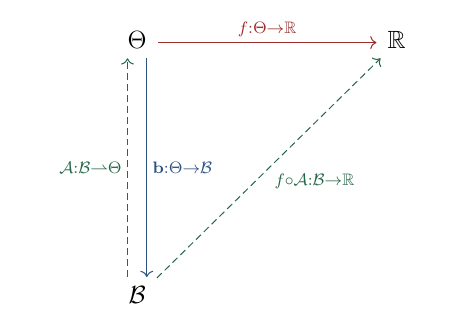}
    \caption{MAP-Elites (Definition~\ref{def:mapelites}). Solid blue: behavioral descriptor $\mathbf{b} : \Theta \to \mathcal{B}$; solid red: fitness $f : \Theta \to \mathbb{R}$; dashed green: archive $\mathcal{A} : \mathcal{B} \rightharpoonup \Theta$ and composed map $f \circ \mathcal{A} : \mathcal{B} \to \mathbb{R}$.}
    \label{fig:commutative}
  \end{subfigure}\hfill
  \begin{subfigure}[t]{0.48\textwidth}
    \centering
    \includegraphics[width=\textwidth]{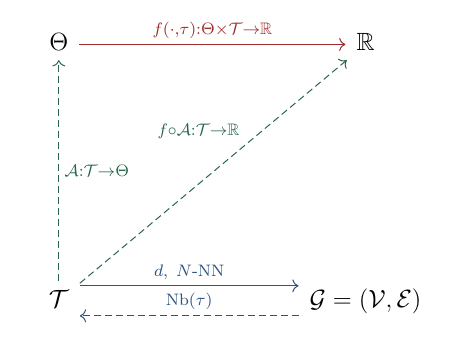}
    \caption{MONET (Definition~\ref{def:monet}). The behavioral descriptor $\mathbf{b}$ is absent; solid red: fitness $f(\cdot,\boldsymbol{\tau}) : \Theta \times \mathcal{T} \to \mathbb{R}$; dashed green: archive $\mathcal{A} : \mathcal{T} \to \Theta$ and $f \circ \mathcal{A} : \mathcal{T} \to \mathbb{R}$; solid blue: graph construction via task distance $d$; dashed blue: neighborhood lookup $\mathrm{Nb}(\boldsymbol{\tau})$.}
    \label{fig:monet_commutative}
  \end{subfigure}
  \caption{Commutative diagrams comparing MAP-Elites and MONET.}
  \label{fig:commutative_combined}
\end{figure}

Figure~\ref{fig:commutative} summarizes the mappings of Definition~\ref{def:mapelites} as a commutative diagram.\footnote{ Note that Definition~\ref{def:mapelites} uses a single-task fitness $f : \Theta \to \mathbb{R}$; the multi-task extensions below lift this to $f : \Theta \times \mathcal{T} \to \mathbb{R}$ by conditioning on a task parameter $\boldsymbol{\tau} \in \mathcal{T}$, recovering the problem statement of Eq.~\eqref{eq:multitask}.}
Now we are ready to introduce relevant multi-task variants as extensions of this definition: each paragraph names exactly what is changed and leaves the rest of the definition in place.

\noindent\textbf{Multi-Task MAP-Elites (MT-ME)~\cite{mouretQualitydiversitymultitaskoptimization2020}.} MT-ME extends Definition~\ref{def:mapelites} by \emph{replacing} the behavioral descriptor $\mathbf{b} : \Theta \to \mathcal{B}$ with the task parameter itself. A finite set of tasks $\{\boldsymbol{\tau}_1, \ldots, \boldsymbol{\tau}_n\} \subset \mathcal{T}$ is fixed in advance, the archive becomes $\mathcal{A} : \mathcal{T} \to \Theta$ indexed directly by task, and the fitness lifts to $f : \Theta \times \mathcal{T} \to \mathbb{R}$, matching the problem statement of Eq.~\eqref{eq:multitask}. Recombination draws parents globally from the entire task archive, so information is shared across all tasks regardless of their similarity. MT-ME uses the iso-line-dd variation operator~\cite{Hypervolume_MAP-Elites}, which exploits the observation that elites tend to cluster in a compact ``elite hypervolume'' in genotype space: given two parent elites $\boldsymbol{\theta}_i$ and $\boldsymbol{\theta}_j$, a candidate is generated by adding isotropic Gaussian noise and a directional component along the line connecting the parents, thereby biasing search toward the region of genotype space where elites concentrate.
Mouret and Maguire~\cite{mouretQualitydiversitymultitaskoptimization2020} evaluated MT-ME on arm and hexapod (Figure \ref{fig:benchmark-suite}), comparing against CMA-ES run independently per task, MAP-Elites with random task assignment, and MAP-Elites evaluated on all tasks simultaneously. MT-ME outperformed all baselines on both domains.
\noindent\textbf{Multi-Task Multi-Behavior MAP-Elites (MTMB-ME)~\cite{MTMB-ME}.} MTMB-ME extends MT-ME by \emph{restoring} the behavioral descriptor from Definition~\ref{def:mapelites} on top of the task index, yielding a two-level archive $\mathcal{A} : \mathcal{T} \times \mathcal{B} \to \Theta$ indexed jointly by task and behavior. Task-specific adaptation and behavioral diversity are therefore maintained simultaneously within a single archive. Anne and Mouret~\cite{MTMB-ME} applied MTMB-ME to a humanoid robot fault-recovery scenario, identifying successful contact positions on a wall that help the robot regain balance after a leg fault, and outperformed random search, grid search, and standard MAP-Elites both in the number of tasks solved and in the number of solutions found.

\noindent\textbf{Parametric-Task MAP-Elites (PT-ME)~\cite{anneParametricTaskMAPElites2024}.} PT-ME extends MT-ME to \emph{continuous} task spaces. Instead of fixing a finite set of tasks in advance, PT-ME draws a fresh task $\boldsymbol{\tau} \sim \mathcal{T}$ at every iteration and stores, for each drawn task, the best solution encountered so far, so the archive becomes a dense dataset covering the entire task space. PT-ME alternates between two variation operators at equal probability. The first is SBX~\cite{SBX} with a bandit-tuned tournament that biases task selection toward tasks close to the parents, replacing the iso-line-dd operator of MT-ME. The second, and more distinctive, operator is a \emph{local linear regression}: PT-ME partitions the task space into cells via Centroidal Voronoi Tessellation (CVT)~\cite{CVT,CVT-MAP-Elites}, precomputes a $k$-dimensional tree ($k$-d tree)~\cite{bentley1975multidimensional} for fast cell lookup and a Delaunay triangulation~\cite{delaunay1934} for adjacency, then fits a linear least-squares model $M = (\mathbf{T}^\top \mathbf{T})^{-1}\mathbf{T}^\top \mathbf{X}$ to the elites in adjacent cells, where $\mathbf{T}$ stacks the task vectors of the adjacent cells and $\mathbf{X}$ the corresponding solution vectors; a candidate is then generated as $\boldsymbol{\theta} = M \cdot \boldsymbol{\tau} + \sigma_{\mathrm{reg}} \cdot \mathcal{N}(0, \mathrm{var}(\mathbf{X}))$. This regression operator is notable because it explicitly exploits task-space locality: the Delaunay triangulation defines a geometric neighborhood over $\mathcal{T}$, and the linear model builds a local approximation of the task-to-solution mapping from that neighborhood. On archery and arm PT-ME yields improved performance over standard MAP-Elites and Proximal Policy Optimization (PPO)~\cite{PPO}.

\noindent\textbf{MAP-Elites with Knowledge Transfer (MMKT)~\cite{9870374}.} This variant extends MT-ME by \emph{partitioning} the task set into groups $\{\mathcal{T}_1, \ldots, \mathcal{T}_m\}$ obtained by applying $k$-means clustering to the task parameters, and then promoting knowledge transfer between tasks through linear combinations of solutions drawn from the same group. In this preliminary study, MMKT outperforms MT-ME on the arm task.

\noindent\textbf{Adaptive Group-based Collaborative Optimization (AGCO)~\cite{houGroupBasedManyTaskCollaborativeOptimizationFrameworkEvolutionaryRobotsDesign2025}.} This framework extends the group-based view further with a two-stage transfer protocol. An \emph{inter-group knowledge separation} stage employs an adaptive selection mechanism for crossover operators to control information exchange between groups, and an \emph{intra-group knowledge reunion} stage leverages Thompson sampling to weight within-group recombination so that target tasks assimilate insights from multiple in-group sources according to their evolutionary preferences.

Most of the variants above treat the archive as fully connected: every elite can serve as a parent for any cell regardless of the relationship between the corresponding tasks. PT-ME is a partial exception: its regression operator builds a local linear model from Delaunay-adjacent cells, thereby exploiting geometric locality in $\mathcal{T}$ for candidate generation. However, this locality is confined to one of two variation operators (applied with 50\% probability), the adjacency structure is a by-product of the CVT tessellation rather than an explicit design element, and the SBX path still draws parents globally. An alternative that makes locality a first-class principle is familiar from \emph{cellular evolutionary algorithms} (cEAs)~\cite{alba2008cellular}, which organize the population into local neighborhoods and restrict recombination accordingly, so that information propagates through the population only via those neighborhoods. MONET (Section~\ref{lab:methods}) follows the same intuition but imposes the graph over tasks rather than over individuals: it extends MT-ME by replacing the implicit fully-connected archive $\mathcal{A} : \mathcal{T} \to \Theta$ with an explicit graph structure over $\mathcal{T}$, inducing neighborhoods from task-space proximity.

\section{MONET}\label{lab:methods}

Building on the taxonomy introduced in Section~\ref{lab:related}, we now introduce MONET. Figure~\ref{fig:teaser} illustrates the task-graph representation at the core of MONET: nodes are tasks, edges connect neighboring tasks in the task parameter space, and each node stores the best solution found so far.

\begin{definition}[MONET]\label{def:monet}
Let $\{\boldsymbol{\tau}_1,\ldots,\boldsymbol{\tau}_n\}\subset\mathcal{T}$ be a finite task set, $f:\Theta\times\mathcal{T}\to\mathbb{R}$ a task-parameterized fitness function, and $d:\mathcal{T}\times\mathcal{T}\to\mathbb{R}_{\ge 0}$ a task distance. MONET represents the optimization state as a labeled graph $\mathcal{G}=(\mathcal{V},\mathcal{E},\mathcal{A})$ whose nodes $\mathcal{V}=\{\boldsymbol{\tau}_1,\ldots,\boldsymbol{\tau}_n\}$ are the tasks themselves, whose edges $\mathcal{E}=\{(\boldsymbol{\tau}_i,\boldsymbol{\tau}_j):\boldsymbol{\tau}_j\text{ is among the }N \text{ nearest neighbors of } \boldsymbol{\tau}_i\text{ under }d\}$ encode task-space proximity for a user-chosen neighborhood size $N$, and whose labeling $\mathcal{A}:\mathcal{V}\to\Theta$ stores the best solution found so far for each task. Candidate solutions are admitted by the elitist rule: a candidate $\boldsymbol{\theta}$ generated for task $\boldsymbol{\tau}_i$ replaces $\mathcal{A}[\boldsymbol{\tau}_i]$ iff $f(\boldsymbol{\theta},\boldsymbol{\tau}_i) \ge f(\mathcal{A}[\boldsymbol{\tau}_i],\boldsymbol{\tau}_i)$.
\end{definition}

Figure~\ref{fig:monet_commutative} summarizes the mappings of Definition~\ref{def:monet}. Compared with the base MAP-Elites diagram (Figure~\ref{fig:commutative}), the behavioral descriptor $\mathbf{b}$ is absent and the archive is indexed directly by task; the fitness is conditioned on a task parameter; and the bottom row shows the graph construction from the task distance $d$ and the neighborhood lookup $\mathrm{Nb}(\boldsymbol{\tau})$ that mediates social learning.

Candidates are generated by one of two complementary mechanisms, selected at each step with probability $p_{\text{ind}}$. \emph{Individual learning} mutates the current label $\mathcal{A}[\boldsymbol{\tau}_i]$ of a focal task, exploring the local fitness landscape of $\boldsymbol{\tau}_i$ in isolation. \emph{Social learning} draws a neighbor $\boldsymbol{\tau}_j$ from the graph neighborhood of $\boldsymbol{\tau}_i$ and recombines $\mathcal{A}[\boldsymbol{\tau}_i]$ with $\mathcal{A}[\boldsymbol{\tau}_j]$, transferring information between similar tasks along the edges of $\mathcal{G}$. Empty nodes are initialized by individual learning with a uniform random solution, or by social learning by copying from a non-empty neighbor (with a uniform random fallback if no such neighbor exists); social learning is skipped at a step where the chosen neighbor is itself empty.

In this work, we use the Euclidean distance in $\mathcal{T}$ as the task distance $d$, and set the neighborhood size $N$ (the number of nearest neighbors retained per task, Def.~\ref{def:monet}) to a small fraction of $|\mathcal{V}|$ (see Section~\ref{lab:experimental}), so that $\mathcal{G}$ is a sparse, partially connected graph whose density is controlled by the chosen neighborhood size. Algorithm~\ref{alg:monet} provides the pseudocode of MONET and the code is publicly accessible~\cite{monet_code}.

\begin{algorithm}[!t]
\caption{MONET}\label{alg:monet}
\footnotesize
\setlength{\textfloatsep}{6pt plus 1pt minus 1pt}
\begin{algorithmic}[1]
\Require Task set $T=\{\boldsymbol{\tau}_1,\dots,\boldsymbol{\tau}_n\}$, fitness function $f$, task distance $d$
\Require Budget $E$, init fraction $K_{\text{init}}$, neighborhood size $N$, selection strategy \Call{Select}{$\cdot$}, probability $p_{\text{ind}}$
\Ensure Archive $\mathcal{A}$ mapping each task to its best solution
\State $\mathcal{A}[\boldsymbol{\tau}] \gets \emptyset$ for all $\boldsymbol{\tau} \in T$ \Comment{empty network}
\For{each $\boldsymbol{\tau}_i \in T$} \Comment{build $k$-NN task graph}
    \State $\mathrm{Nb}(\boldsymbol{\tau}_i) \gets N$ nearest tasks under $d$
\EndFor
\For{$i \gets 1$ \textbf{to} $K_{\text{init}}$} \Comment{random initialization}
    \State $\boldsymbol{\tau} \gets \Call{RandomTask}{T}$
    \State $\boldsymbol{\theta} \sim \mathcal{U}(p_{\min}, p_{\max})$
    \State $f_x \gets f(\boldsymbol{\theta}, \boldsymbol{\tau})$
    \If{$\mathcal{A}[\boldsymbol{\tau}] = \emptyset$ \textbf{or} $f_x \geq \mathcal{A}[\boldsymbol{\tau}].\text{fitness}$} \Comment{elitist update}
        \State $\mathcal{A}[\boldsymbol{\tau}] \gets (\boldsymbol{\theta}, f_x)$
    \EndIf
\EndFor
\State $e \gets K_{\text{init}}$ \Comment{main loop}
\While{$e < E$}
    \State $\boldsymbol{\tau} \gets \Call{RandomTask}{T}$ \Comment{sample focal task}
    \If{$\mathrm{rand}() < p_{\text{ind}}$} \Comment{individual learning}
        \State \Call{Mutation}{$\boldsymbol{\tau}, \mathcal{A}$} \Comment{\Call{IndividualLearning}{$\boldsymbol{\tau}, \mathcal{A}$}}
        \State $e \gets e + 1$
    \Else \Comment{social learning}
        \State $\boldsymbol{\tau}_j \gets \Call{Select}{\mathrm{Nb}(\boldsymbol{\tau})}$ \Comment{pick one neighbor}
        \If{$\boldsymbol{\tau}_j \neq \text{nil}$} \Comment{skip if no graph neighbor exists}
            \State \Call{SBX}{$\boldsymbol{\tau}, \boldsymbol{\tau}_j, \mathcal{A}$} \Comment{\Call{SocialLearning}{$\boldsymbol{\tau}, \boldsymbol{\tau}_j, \mathcal{A}$}}
            \State $e \gets e + 1$
        \EndIf
    \EndIf
\EndWhile
\State \Return $\mathcal{A}$
\end{algorithmic}
\end{algorithm}









\section{Experimental Setup}\label{lab:experimental}

\begin{figure}[t]
    \centering
    \makebox[\textwidth]{\includegraphics[width=1\textwidth]{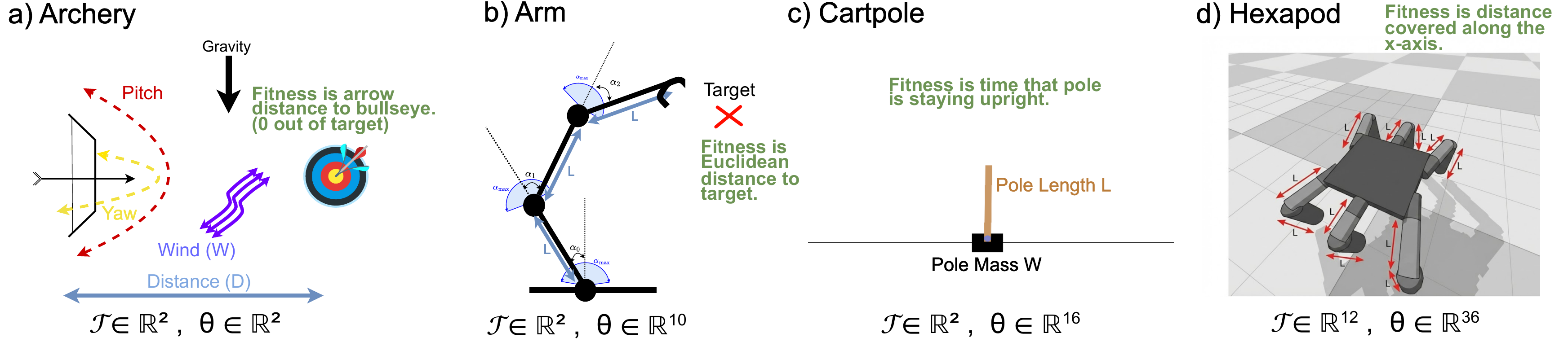}}
    \caption{Overview of benchmark environments.}
    \label{fig:benchmark-suite}
\end{figure}

We compare MONET against PT-ME and MT-ME across four task environments: archery, arm, cartpole, and hexapod (Figure \ref{fig:benchmark-suite}).

PT-ME samples tasks continuously from $\mathcal{T}$ rather than operating over a fixed task set: it partitions $\mathcal{T}$ into $n$ cells via Centroidal Voronoi Tessellation (CVT)~\cite{CVT,CVT-MAP-Elites} and, by the end of its run, each cell holds the highest-fitness task-solution pair it has sampled. To place the three algorithms on the same footing, we take these $n$ cell elites ($n=5000$ for archery, arm, and cartpole, and $n=2000$ for hexapod) as the fixed task set on which MT-ME and MONET are then run. This removes task selection as a confound in the cross-algorithm comparison. One asymmetry remains: under this protocol each PT-ME task is evaluated essentially once (the sample whose solution became its cell's elite), whereas MT-ME and MONET revisit each task many times to refine its solution, so the total evaluation budget is identical across algorithms but the per-task refinement opportunity is not.

For PT-ME and MT-ME we use the default parameters from \cite{anneParametricTaskMAPElites2024,mouretQualitydiversitymultitaskoptimization2020}.
For MONET, we first conduct a hyperparameter sensitivity analysis.
We consider the following hyperparameters and ranges: $p_\text{ind} \in \{0,\, 0.3,\, 0.5,\, 0.7,\, 1\}$, $\text{strategy} \in \{\text{random},\, \text{best\_fitness}\}$, $\text{neighborhood} \in \{\text{random},\, \text{closest},\, \text{distance\_prop.}\}$, and $\text{neighbor\_percentage} \in \{1\%,\, 5\%,\, 10\%\}$. Here $p_\text{ind}$ is the per-step probability of performing individual learning (mutation) rather than social learning (SBX); $\text{strategy}$ is the rule used to pick the neighbor during social learning, either uniformly at random or deterministically the neighbor with the highest current fitness; $\text{neighborhood}$ determines how the neighbor set of each task is built, taking the $N$ closest tasks in $\mathcal{T}$, $N$ uniformly random tasks, or $N$ tasks drawn without replacement with probability proportional to task-space similarity; and $\text{neighbor\_percentage}$ sets this neighborhood size as a fraction $N = \lceil \text{neighbor\_percentage} \cdot |\mathcal{V}| \rceil$ of the total task set.

For each hyperparameter combination we perform up to 20 independent runs with different random seeds and report the mean fitness with standard deviation as well as the area under the learning curve (AUC), capturing both final performance and cumulative learning efficiency. The sensitivity sweep is concentrated on archery and arm (20 seeds per configuration; cartpole and hexapod are evaluated on a smaller subset of configurations because of simulator cost).

To quantify the importance of individual hyperparameters and their interactions, both per task and aggregated across all tasks, we fit a gradient-boosted tree surrogate model and compute SHapley Additive exPlanations (SHAP) values~\cite{lundberg2017unified,lundberg2020local2global}, following the surrogate-based hyperparameter importance methodology of~\cite{hutter2014efficient}.
Guided by this analysis, we then assign reasonable values to each hyperparameter, yielding a single configuration of MONET that we refer to as the \emph{default} configuration and compare it against PT-ME and MT-ME.
Additionally, we employ the Sequential Parameter Optimization Toolbox (SPOT)~\cite{bartzbeielstein2026optimizationspotoptim}, a surrogate-based optimizer, to automatically tune MONET's hyperparameters per task and include these task-specific configurations in the comparison.

To assess statistical significance of the performance differences between algorithms, we apply the Mann-Whitney U test~\cite{mann1947test} to the final fitness values and AUC values across all seeds and report the resulting $p$-values with Holm-Bonferroni correction~\cite{holm1979simple,dunn1961multiple} for multiple comparisons.

Higher fitness is better in all four benchmarks, but the scale differs across them: archery and arm return values in $[0,1]$; cartpole reports the mean balanced-step count across 10 rollouts and therefore lies in $[0, t_{\max}]$ with $t_{\max}=1000$; hexapod reports the forward centre-of-mass displacement in metres over a 3\,s episode with no formal upper bound (in practice below $1.5\,\mathrm{m}$). Closed-form fitness expressions are defined in the Supplementary Information (SI)~\cite{monet_code}. All four benchmarks are run for $10^6$ evaluations and 20 seeds per run.

\paragraph{Archery.} A projectile-aiming task in which the archer compensates for target distance and wind (Figure~\ref{fig:benchmark-suite}a)~\cite{anneParametricTaskMAPElites2024}. Tasks (2D) vary the distance $D\in[5,40]$\,m and horizontal wind $W\in[-10,10]\,\mathrm{m/s^2}$; solutions (2D) are yaw and pitch in $[-\pi/12,\pi/12]$ at fixed launch speed $v_0=70\,\mathrm{m/s}$.

\paragraph{Arm.} A planar 10-degree-of-freedom (10-DoF) arm places its end-effector at a fixed target (Figure~\ref{fig:benchmark-suite}b)~\cite{mouretQualitydiversitymultitaskoptimization2020,anneParametricTaskMAPElites2024}. Tasks (2D) vary the link length $L$ and maximum joint angle $\alpha_{\max}$, both in $[0,1]$; solutions (10D) are the 10 joint angles.

\paragraph{Cartpole.} Cart-pole balancing under varying physical parameters (Figure~\ref{fig:benchmark-suite}c)~\cite{michie1968boxes,barto1983neuronlike}. Tasks (2D) vary pole mass $m_p\in[0.05,0.5]$\,kg and pole length $L\in[0.3,1.2]$\,m; solutions (16D) are a direct encoding of the weights of a small multi-layer perceptron controller (4 inputs, 2 hidden tanh units, 2 output logits, with bias units on the input and hidden layers).

\paragraph{Hexapod.} A simulated hexapod walks forward across 2\,000 random morphologies (Figure~\ref{fig:benchmark-suite}d)~\cite{MAP_Elites_NaturePaper,mouretQualitydiversitymultitaskoptimization2020}, simulated in PyBullet~\cite{coumans2021pybullet}. Tasks (12D) are 12 leg-segment length changes; solutions (36D) parameterise a gait controller.

\section{Results}\label{lab:results}

Across the four domains, MONET matches or exceeds the strongest MAP-Elites baseline on final mean fitness, with the largest gap on the hexapod task where the per-domain tuned configuration nearly doubles the final fitness of MT-ME (Figure~\ref{fig:algorithm_comparison}, Table~\ref{tab:results-summary}). We first walk through the per-domain comparison, then unpack the hyperparameter analysis that produced the shared default configuration.

\subsection{Algorithm Comparison}\label{sec:comparison}

\begin{figure}[H]
    \centering
    \includegraphics[width=\textwidth]{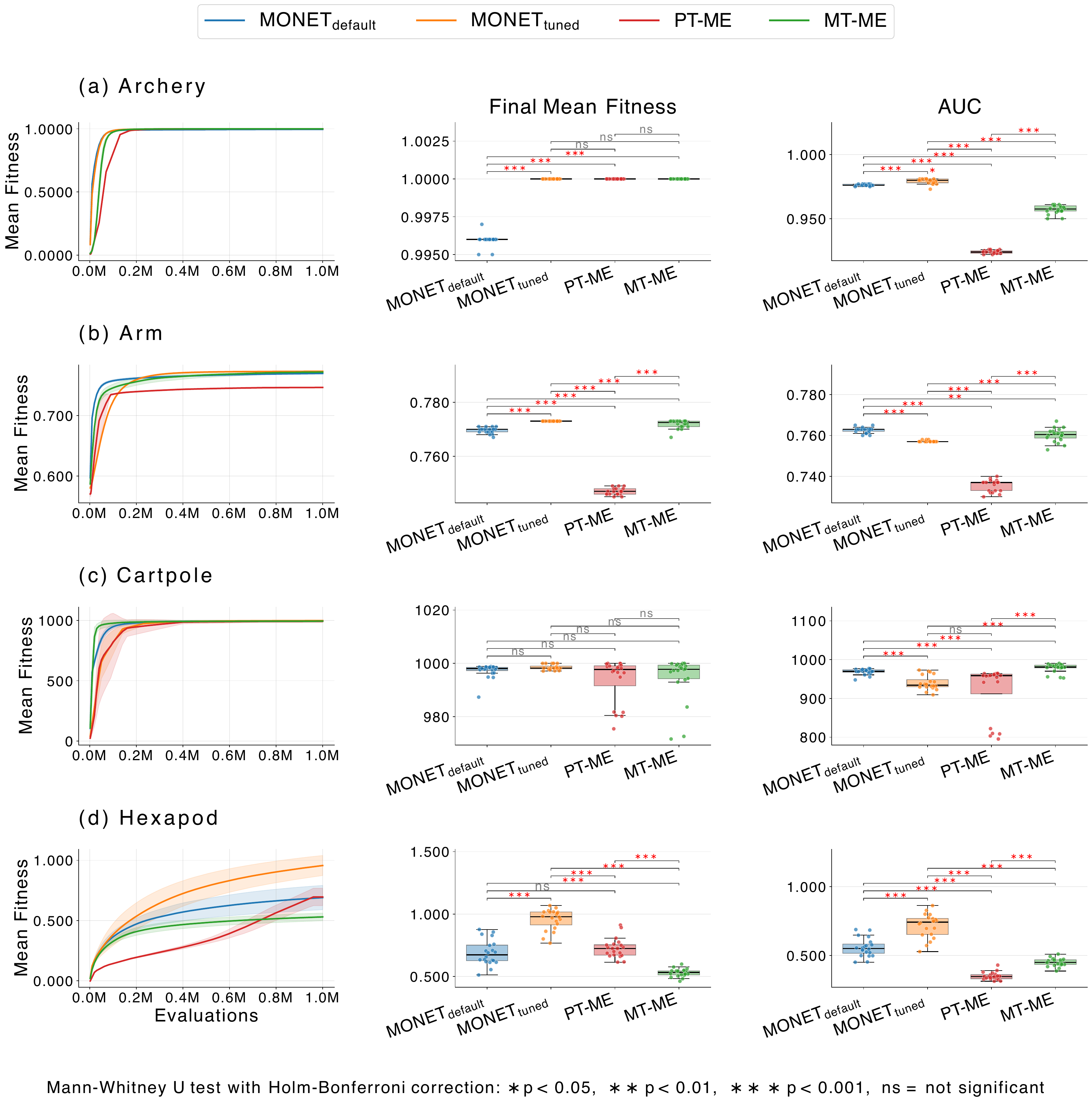}
    \caption{Algorithm comparison across the four benchmark domains.}
    \label{fig:algorithm_comparison}
\end{figure}

We report two MONET variants: \emph{MONET$_{\mathrm{default}}$}, a single configuration shared across domains (defined in Section~\ref{sec:default-config}), and \emph{MONET$_{\mathrm{tuned}}$}, a per-domain tuning produced with SPOT \cite{bartzbeielstein2026optimizationspotoptim}. Both are compared against PT-ME and MT-ME with their published defaults. Each box in Figure~\ref{fig:algorithm_comparison} aggregates 20 seeds per (algorithm, domain) pair; brackets indicate pairwise comparisons where one algorithm is significantly better (Mann-Whitney U, Holm-Bonferroni correction, $p<0.05$).

\paragraph{Archery.} MT-ME, PT-ME, and MONET$_{\mathrm{tuned}}$ all saturate at $1.000$; MONET$_{\mathrm{default}}$ stops short at $0.996 \pm 0.000$ ($p<0.001$ vs.\ all three). Final fitness is thus tied at the top, and AUC is the discriminating signal. MONET$_{\mathrm{default}}$ reaches an AUC of $976.3 \pm 0.6$, significantly above MT-ME ($957.2 \pm 3.3$, $p<0.001$) and PT-ME ($924.1 \pm 1.1$, $p<0.001$). MONET$_{\mathrm{tuned}}$ improves AUC further to $979.2 \pm 1.9$ ($p<0.001$ vs.\ MONET$_{\mathrm{default}}$). MONET therefore converges to near-optimal performance substantially earlier than the baselines.

\paragraph{Arm.} MONET$_{\mathrm{default}}$ reaches $0.770 \pm 0.001$ on final fitness, just below MT-ME ($0.772 \pm 0.002$, $p<0.001$) and above PT-ME ($0.747 \pm 0.001$, $p<0.001$). On AUC it is the strongest of the four algorithms: $762.7 \pm 1.2$, ahead of MT-ME ($760.1 \pm 3.4$, $p=0.003$) and PT-ME ($734.0 \pm 2.9$, $p<0.001$). This reflects faster convergence in the early phase (Figure~\ref{fig:algorithm_comparison}b). MONET$_{\mathrm{tuned}}$ achieves the highest final fitness at $0.773 \pm 0.000$, significantly above MT-ME ($p<0.001$).

\paragraph{Cartpole.} MONET$_{\mathrm{default}}$ reaches $997.2 \pm 2.6$, statistically indistinguishable from MT-ME ($994.5 \pm 8.5$) and PT-ME ($992.5 \pm 8.9$). MT-ME retains a significant AUC lead ($978.4 \pm 11.4$ vs.\ $969.0 \pm 7.2$, $p<0.001$): it converges earlier while MONET terminates higher. Tuning nudges final fitness to $998.5 \pm 1.0$ (not significant vs.\ baselines) but significantly lowers AUC to $941.4 \pm 17.4$ ($p<0.001$ vs.\ MONET$_{\mathrm{default}}$).

\paragraph{Hexapod.} MONET$_{\mathrm{default}}$ ($0.690 \pm 0.101$) performs significantly better than MT-ME ($0.530 \pm 0.032$, $p<0.001$) and statistically indistinguishable from PT-ME ($0.727 \pm 0.078$, $p=0.15$). Tuning improves performance further to $0.957 \pm 0.083$, significantly above both baselines ($p<0.001$ vs.\ MT-ME; $p<0.001$ vs.\ PT-ME).

Previous work reported that PT-ME outperformed MT-ME~\cite{anneParametricTaskMAPElites2024}, whereas in our experiments, MT-ME achieved better performance than PT-ME. This discrepancy arises because PT-ME samples tasks continuously from the task space, which can result in selecting tasks that are easier to solve. To ensure a fair comparison, we fixed the task set to the final tasks discovered by PT-ME and evaluated both methods on this same set of tasks.

\begin{table}[H]
\centering
\caption{Final mean fitness and AUC ($\times 10^3$ on the $[0,1]$ domains) per domain. Mean $\pm$ std over 20 random seeds per (algorithm, domain) pair.}
\label{tab:results-summary}
\small
\setlength{\tabcolsep}{3pt}
\begin{tabular}{@{}lcccc@{}}
\toprule
 & Archery & Arm & Cartpole & Hexapod \\
\midrule
\multicolumn{5}{@{}l}{\textit{(a) Final mean fitness}} \\[2pt]
PT-ME             & $1.000 \pm 0.000$         & $0.747 \pm 0.001$         & $992.5 \pm 8.9$           & $0.727 \pm 0.078$ \\
MT-ME             & $1.000 \pm 0.000$         & $0.772 \pm 0.002$         & $994.5 \pm 8.5$           & $0.530 \pm 0.032$ \\
MONET$_{\mathrm{default}}$   & $0.996 \pm 0.000$         & $0.770 \pm 0.001$         & $997.2 \pm 2.6$           & $0.690 \pm 0.101$ \\
MONET$_{\mathrm{tuned}}$      & $1.000 \pm 0.000$         & $\mathbf{0.773 \pm 0.000}$ & $998.5 \pm 1.0$           & $\mathbf{0.957 \pm 0.083}$ \\
\midrule
\multicolumn{5}{@{}l}{\textit{(b) AUC ($\times 10^3$ on $[0,1]$ domains)}} \\[2pt]
PT-ME             & $924.1 \pm 1.1$           & $734.0 \pm 2.9$           & $913.5 \pm 70.2$          & $349.9 \pm 27.7$  \\
MT-ME             & $957.2 \pm 3.3$           & $760.1 \pm 3.4$           & $\mathbf{978.4 \pm 11.4}$ & $450.2 \pm 29.3$  \\
MONET$_{\mathrm{default}}$   & $976.3 \pm 0.6$           & $\mathbf{762.7 \pm 1.2}$  & $969.0 \pm 7.2$           & $557.0 \pm 67.3$  \\
MONET$_{\mathrm{tuned}}$      & $\mathbf{979.2 \pm 1.9}$  & $757.3 \pm 0.2$           & $941.4 \pm 17.4$          & $\mathbf{716.0 \pm 87.8}$ \\
\bottomrule
\end{tabular}
\end{table}

\subsection{Hyperparameter Sensitivity and Default Configuration}\label{sec:default-config}\label{sec:sensitivity}

\begin{figure}[H]
    \centering
    \includegraphics[width=0.7\linewidth]{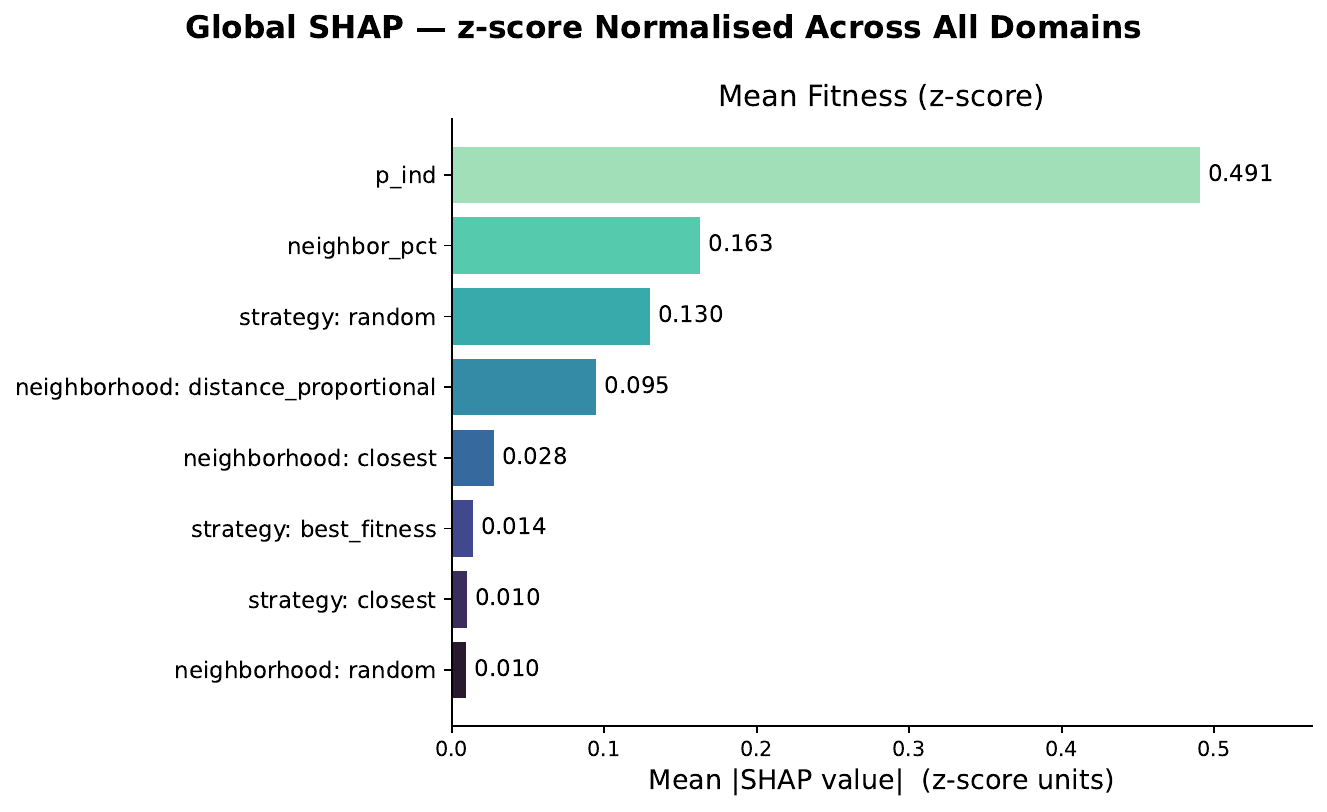}
    \caption{Global hyperparameter importance for MONET: mean absolute SHAP value on a gradient-boosted tree surrogate.}
    \label{fig:shap-global}
\end{figure}

The SHAP analysis identifies $p_\text{ind}$ as the dominant hyperparameter ($\overline{|\text{SHAP}|}=0.49$), with low values (favoring social over individual learning) consistently improving fitness and AUC across all domains. The remaining parameters show weaker, domain-dependent effects (Figure~\ref{fig:shap-global}; see SI~\cite{monet_code}).

Based on this analysis, we identify the default configuration as: strategy $=$ best\_fitness, neighborhood$=$closest, $p_\text{ind}=0.3$, neighbor\_percentage$=5\%$. SPOT-tuned per-domain configurations differ markedly, indicating that the optimal individual--social learning balance is task-dependent.

\section{Conclusion}\label{lab:discussion}

In this work, we propose MONET to find high-quality solutions for a large set of related tasks simultaneously and in parallel. We propose structuring the task space as a network to establish topological relations and neighborhoods, which aims to exploit task similarities for knowledge transfer. As shown by the results in four domains, this approach has an advantage over flat archive representation, where these topological relations are ignored. 
We employ two operators, namely individual and social learning, that aim to improve tasks in isolation or exploit knowledge transfer across neighboring related tasks. We chose these names to refer to a wide range of individual or social learning strategies that are not limited to mutation and crossover. For instance, among others, we aim to use Evolution Strategies for individual and direct solution copying for social learning in future studies.

The extensive sensitivity analysis of the MONET algorithm and its parameters showed that the probability of individual learning $p_\text{ind}$ influences the performance the most, with high values leading to lower performance. This shows that
social learning is a significant contributor to performance by exploiting knowledge transfer across tasks. However, this analysis was on a pre-selected set of $p_\text{ind}$ parameters. Thus, using a more balanced ratio between individual and social learning parameters can improve performance, particularly in task spaces
where similarities between tasks are relatively low. In fact, SPOT-based per-domain tuning selected low $p_\text{ind}$ on hexapod $(0)$ and archery $(0.22)$. In arm $(0.54)$ and cartpole $(0.5)$, a more balanced mix was preferred.

We establish connectivity between tasks on the basis of the task parameter similarities. This relies on the idea that tasks sharing similar parameters are themselves similar, making them more likely to benefit from knowledge transfer and to converge toward similar solutions. 

However, this assumption may not hold in many domains. In the hexapod domain, for example, small changes to front-right and back-left leg lengths yield nearby task vectors but can demand qualitatively different solutions to perform the control tasks. Future work should therefore explore alternative connectivity definitions and assess their effect on knowledge transfer.

Currently, the MONET algorithm has four hyperparameters. Another interesting direction is online self-adaptive parameter control (e.g., adapting the neighborhood size as in MT-ME's task-competition and replacement rule~\cite{mouretQualitydiversitymultitaskoptimization2020}, or dynamically switching between individual and social learning strategies~\cite{yaman2022meta}). Although this can introduce complexity, it may provide more generalized performance across different domains and task spaces.
Another promising direction is to extend the types of individual and social
learning strategies that can improve performance. Furthermore, some of these
strategies can converge to more general solutions that can lead to satisfactory performance for a wide range of tasks in the task space ~\cite{triebold2024evolving}.



\bibliographystyle{splncs04}
\bibliography{refs}

@book{deb2001multi,
  title = {Multi-Objective Optimization Using Evolutionary Algorithms},
  author = {Deb, Kalyanmoy},
  year = {2001},
  publisher = {Wiley},
  address = {Chichester, UK},
  isbn = {978-0-471-87339-6}
}

@book{feller1968introduction,
  title = {An Introduction to Probability Theory and Its Applications},
  author = {Feller, William},
  volume = {1},
  edition = {3},
  year = {1968},
  publisher = {Wiley}
}

@misc{coumans2021pybullet,
  title = {{PyBullet}, a {Python} module for physics simulation for games, robotics and machine learning},
  author = {Coumans, Erwin and Bai, Yunfei},
  year = {2021},
  howpublished = {\url{http://pybullet.org}}
}

@incollection{michie1968boxes,
  title = {{BOXES}: An Experiment in Adaptive Control},
  author = {Michie, Donald and Chambers, Roger A.},
  booktitle = {Machine Intelligence 2},
  editor = {Dale, E. and Michie, D.},
  pages = {137--152},
  year = {1968},
  publisher = {Oliver and Boyd},
  address = {Edinburgh}
}

@article{barto1983neuronlike,
  title = {Neuronlike Adaptive Elements That Can Solve Difficult Learning Control Problems},
  author = {Barto, Andrew G. and Sutton, Richard S. and Anderson, Charles W.},
  journal = {IEEE Transactions on Systems, Man, and Cybernetics},
  volume = {SMC-13},
  number = {5},
  pages = {834--846},
  year = {1983},
  doi = {10.1109/TSMC.1983.6313077}
}

@article{yaman2021distributed,
  title={Distributed embodied evolution over networks},
  author={Yaman, Anil and Iacca, Giovanni},
  journal={Applied Soft Computing},
  volume={101},
  pages={106993},
  year={2021},
  publisher={Elsevier}
}

@article{gupta2015multifactorial,
  title = {Multifactorial Evolution: Toward Evolutionary Multitasking},
  author = {Gupta, Abhishek and Ong, Yew-Soon and Feng, Liang},
  journal = {IEEE Transactions on Evolutionary Computation},
  volume = {20},
  number = {3},
  pages = {343--357},
  year = {2016},
  doi = {10.1109/TEVC.2015.2458037}
}

@inproceedings{anneParametricTaskMAPElites2024,
  title = {Parametric-{{Task MAP-Elites}}},
  booktitle = {Proceedings of the {{Genetic}} and {{Evolutionary Computation Conference}}},
  author = {Anne, Timoth{\'e}e and Mouret, Jean-Baptiste},
  year = 2024,
  month = jul,
  eprint = {2402.01275},
  primaryclass = {cs},
  pages = {68--77},
  doi = {10.1145/3638529.3653993},
  urldate = {2025-06-02},
  archiveprefix = {arXiv},
  langid = {english}
}

@article{Bayesian-Optimization_review,
  title = {Taking the Human out of the Loop: A Review of Bayesian Optimization},
  author = {Shahriari, Bobak and Swersky, Kevin and Wang, Ziyu and Adams, Ryan P. and {de Freitas}, Nando},
  year = 2016,
  journal = {Proceedings of the IEEE},
  volume = {104},
  number = {1},
  pages = {148--175},
  doi = {10.1109/JPROC.2015.2494218}
}

@article{CVT,
  title = {Centroidal {{Voronoi}} Tessellations: {{Applications}} and Algorithms},
  author = {Du, Qiang and Faber, Vance and Gunzburger, Max},
  year = 1999,
  journal = {SIAM Review},
  volume = {41},
  number = {4},
  eprint = {https://doi.org/10.1137/S0036144599352836},
  pages = {637--676},
  doi = {10.1137/S0036144599352836}
}

@article{CVT-MAP-Elites,
  author       = {Vassilis Vassiliades and
                  Konstantinos I. Chatzilygeroudis and
                  Jean{-}Baptiste Mouret},
  title        = {Scaling Up MAP-Elites Using Centroidal {{Voronoi}} Tessellations},
  journal      = {CoRR},
  volume       = {abs/1610.05729},
  year         = {2016},
  url          = {http://arxiv.org/abs/1610.05729},
  eprinttype   = {arXiv},
  eprint       = {1610.05729},
  bibsource    = {dblp computer science bibliography, https://dblp.org}
}

@article{delaunay1934,
  author  = {Delaunay, Boris},
  title   = {Sur la sph{\`e}re vide},
  journal = {Bulletin de l'Acad{\'e}mie des Sciences de l'URSS,
             Classe des sciences math{\'e}matiques et naturelles},
  volume  = {6},
  pages   = {793--800},
  year    = {1934}
}

@article{EMT_6_applications,
  title = {Half a Dozen Real-World Applications of Evolutionary Multitasking, and More},
  author = {Gupta, Abhishek and Zhou, Lei and Ong, Yew-Soon and Chen, Zefeng and Hou, Yaqing},
  year = 2022,
  month = may,
  journal = {IEEE Computational Intelligence Magazine},
  volume = {17},
  pages = {49--66},
  doi = {10.1109/MCI.2022.3155332}
}

@article{EMT_complex_design_electric_power_dispatch,
  title = {A Multitasking Electric Power Dispatch Approach with Multi-Objective Multifactorial Optimization Algorithm},
  author = {Liu, Junwei and Li, Peiling and Wang, Guibin and Zha, Yongxing and Peng, Jianchun and Xu, Gang},
  year = 2020,
  journal = {IEEE access : practical innovations, open solutions},
  volume = {8},
  pages = {155902--155911},
  doi = {10.1109/ACCESS.2020.3018484}
}

@article{EMT_complex_design_photovoltaic_models,
  title = {Evolutionary Multi-Task Optimization for Parameters Extraction of Photovoltaic Models},
  author = {Liang, Jing and Qiao, Kangjia and Yuan, Minghua and Yu, Kunjie and Qu, Boyang and Ge, Shilei and Li, Yaxin and Chen, Guanlin},
  year = 2020,
  month = mar,
  journal = {Energy Conversion and Management},
  volume = {207},
  pages = {112509},
  doi = {10.1016/j.enconman.2020.112509}
}

@article{EMT_logistics_planning_combinatorial_optimzation,
  title = {Explicit Evolutionary Multitasking for Combinatorial Optimization: A Case Study on Capacitated Vehicle Routing Problem},
  author = {Feng, Liang and Huang, Yuxiao and Zhou, Lei and Zhong, Jinghui and Gupta, Abhishek and Tang, Ke and Tan, Kay Chen},
  year = 2021,
  journal = {IEEE Transactions on Cybernetics},
  volume = {51},
  number = {6},
  pages = {3143--3156},
  doi = {10.1109/TCYB.2019.2962865}
}

@inproceedings{EMT_software_testing_multifactorial,
  title = {Concurrently Searching Branches in Software Tests Generation through Multitask Evolution},
  booktitle = {2016 {{IEEE}} Symposium Series on Computational Intelligence ({{SSCI}})},
  author = {Sagarna, Ramon and Ong, Yew-Soon},
  year = 2016,
  pages = {1--8},
  doi = {10.1109/SSCI.2016.7850040}
}

@article{EMT_symbolic_regression_with_GP,
  title = {Multifactorial Genetic Programming for Symbolic Regression Problems},
  author = {Zhong, Jinghui and Feng, Liang and Cai, Wentong and Ong, Yew-Soon},
  year = 2020,
  journal = {IEEE Transactions on Systems, Man, and Cybernetics: Systems},
  volume = {50},
  number = {11},
  pages = {4492--4505},
  doi = {10.1109/TSMC.2018.2853719}
}

@article{EMT_UAV_path_planning,
  title = {Cognizant Multitasking in Multiobjective Multifactorial Evolution: {{MO-MFEA-II}}},
  author = {Bali, Kavitesh Kumar and Gupta, Abhishek and Ong, Yew-Soon and Tan, Puay Siew},
  year = 2021,
  journal = {IEEE Transactions on Cybernetics},
  volume = {51},
  number = {4},
  pages = {1784--1796},
  doi = {10.1109/TCYB.2020.2981733}
}

@article{houGroupBasedManyTaskCollaborativeOptimizationFrameworkEvolutionaryRobotsDesign2025,
  title = {A {{Group-Based Many-Task Collaborative Optimization Framework}} for {{Evolutionary Robots Design}}},
  author = {Hou, Yaqing and Yu, Zhaoping and Guo, Zheng and Pei, Wenbin and Wu, Yaoxin and Ge, Hongwei and Xue, Bing and Zhang, Mengjie},
  year = 2025,
  month = may,
  journal = {IEEE Transactions on Systems, Man, and Cybernetics: Systems},
  volume = {55},
  number = {5},
  pages = {3492--3505},
  issn = {2168-2216, 2168-2232},
  doi = {10.1109/TSMC.2025.3541002},
  urldate = {2026-04-09},
  langid = {english}
}

@article{huynhthithanhEnsembleMultifactorialEvolutionBiasedSkillFactorInheritanceManyTaskOptimization2023,
  title = {Ensemble {{Multifactorial Evolution With Biased Skill-Factor Inheritance}} for {{Many-Task Optimization}}},
  author = {Huynh Thi Thanh, Binh and Van Cuong, Le and Thang, Ta Bao and Long, Nguyen Hoang},
  year = 2023,
  month = dec,
  journal = {IEEE Transactions on Evolutionary Computation},
  volume = {27},
  number = {6},
  pages = {1735--1749},
  issn = {1089-778X, 1089-778X, 1941-0026},
  doi = {10.1109/TEVC.2022.3227120},
  urldate = {2026-04-09},
  langid = {english}
}

@article{Hypervolume_MAP-Elites,
  author       = {Vassilis Vassiliades and
                  Jean{-}Baptiste Mouret},
  title        = {Discovering the Elite Hypervolume by Leveraging Interspecies Correlation},
  journal      = {CoRR},
  volume       = {abs/1804.03906},
  year         = {2018},
  url          = {http://arxiv.org/abs/1804.03906},
  eprinttype   = {arXiv},
  eprint       = {1804.03906},
  bibsource    = {dblp computer science bibliography, https://dblp.org}
}

@article{yaman2022meta,
  title={Meta-control of social learning strategies},
  author={Yaman, Anil and Bredeche, Nicolas and {\c{C}}aylak, Onur and Leibo, Joel Z and Lee, Sang Wan},
  journal={PLoS computational biology},
  volume={18},
  number={2},
  pages={e1009882},
  year={2022},
  publisher={Public Library of Science San Francisco, CA USA}
}

@inproceedings{triebold2024evolving,
  title={Evolving generalist controllers to handle a wide range of morphological variations},
  author={Triebold, Corinna and Yaman, Anil},
  booktitle={Proceedings of the Genetic and Evolutionary Computation Conference},
  pages={1137--1145},
  year={2024}
}

@article{MAP_Elites_NaturePaper,
	author = {Cully, Antoine and Clune, Jeff and Tarapore, Danesh and Mouret, Jean-Baptiste},
	date = {2015/05/01},
	doi = {10.1038/nature14422},
	id = {Cully2015},
	isbn = {1476-4687},
	journal = {Nature},
	number = {7553},
	pages = {503--507},
	title = {Robots that can adapt like animals},
	url = {https://doi.org/10.1038/nature14422},
	volume = {521},
	year = {2015}}

@inproceedings{mouretQualitydiversitymultitaskoptimization2020,
  title = {Quality Diversity for Multi-Task Optimization},
  booktitle = {Proceedings of the 2020 {{Genetic}} and {{Evolutionary Computation Conference}}},
  author = {Mouret, Jean-Baptiste and Maguire, Glenn},
  year = 2020,
  month = jun,
  pages = {121--129},
  publisher = {ACM},
  address = {Canc\'un Mexico},
  doi = {10.1145/3377930.3390203},
  urldate = {2025-06-02},
  isbn = {978-1-4503-7128-5},
  langid = {english}
}

@inproceedings{MTMB-ME,
  title = {Multi-Task Multi-Behavior {{MAP-elites}}},
  booktitle = {Proceedings of the Companion Conference on Genetic and Evolutionary Computation},
  author = {Anne, Timoth{\'e}e and Mouret, Jean-Baptiste},
  year = 2023,
  series = {{{GECCO}} '23 Companion},
  pages = {111--114},
  publisher = {Association for Computing Machinery},
  address = {New York, NY, USA},
  doi = {10.1145/3583133.3590730},
  isbn = {979-8-4007-0120-7}
}

@article{Multi_task_bayesian_optimization,
  title = {Continuous Multi-Task Bayesian Optimisation with Correlation},
  author = {Pearce, Michael and Branke, Juergen},
  year = 2018,
  journal = {European Journal of Operational Research},
  volume = {270},
  number = {3},
  pages = {1074--1085},
  issn = {0377-2217},
  doi = {10.1016/j.ejor.2018.03.017}
}

@article{multi_task_survey,
  title = {What Makes Evolutionary Multi-Task Optimization Better: {{A}} Comprehensive Survey},
  author = {Zhao, Hong and Ning, Xuhui and Liu, Xiaotao and Wang, Chao and Liu, Jing},
  year = 2023,
  journal = {Applied Soft Computing},
  volume = {145},
  pages = {110545},
  issn = {1568-4946},
  doi = {10.1016/j.asoc.2023.110545}
}

@article{PPO,
  author       = {John Schulman and
                  Filip Wolski and
                  Prafulla Dhariwal and
                  Alec Radford and
                  Oleg Klimov},
  title        = {Proximal Policy Optimization Algorithms},
  journal      = {CoRR},
  volume       = {abs/1707.06347},
  year         = {2017},
  url          = {http://arxiv.org/abs/1707.06347},
  eprinttype   = {arXiv},
  eprint       = {1707.06347},
  bibsource    = {dblp computer science bibliography, https://dblp.org}
}

@article{DBLP:journals/corr/MouretC15,
  author       = {Jean{-}Baptiste Mouret and
                  Jeff Clune},
  title        = {Illuminating search spaces by mapping elites},
  journal      = {CoRR},
  volume       = {abs/1504.04909},
  year         = {2015},
  url          = {http://arxiv.org/abs/1504.04909},
  eprinttype    = {arXiv},
  eprint       = {1504.04909},
  bibsource    = {dblp computer science bibliography, https://dblp.org}
}

@article{SBX,
  title = {Simulated Binary Crossover for Continuous Search Space},
  author = {Deb, Kalyanmoy and Agrawal, Ram Bhushan},
  year = 1995,
  journal = {Complex Systems},
  volume = {9},
  number = {2},
  pages = {115--148}
}

@article{wangSolvingMultitaskOptimizationProblemsAdaptiveKnowledgeTransferAnomalyDetection2022,
  title = {Solving {{Multitask Optimization Problems With Adaptive Knowledge Transfer}} via {{Anomaly Detection}}},
  author = {Wang, Chao and Liu, Jing and Wu, Kai and Wu, Zhaoyang},
  year = 2022,
  month = apr,
  journal = {IEEE Transactions on Evolutionary Computation},
  volume = {26},
  number = {2},
  pages = {304--318},
  issn = {1089-778X, 1089-778X, 1941-0026},
  doi = {10.1109/TEVC.2021.3068157},
  urldate = {2026-04-09},
  langid = {english}
}

@INPROCEEDINGS{9870374,
  author={Liu, Peng and Guo, Zheng and Yu, Hua and Linghu, Han and Li, Yijing and Hou, Yaqing and Ge, Hongwei and Zhang, Qiang},
  booktitle={2022 IEEE Congress on Evolutionary Computation (CEC)}, 
  title={A Preliminary Study of Multi-task MAP-Elites with Knowledge Transfer for Robotic Arm Design}, 
  year={2022},
  volume={},
  number={},
  pages={1-8},
  doi={10.1109/CEC55065.2022.9870374}}

@misc{osaba2021evolutionarymultitaskoptimizationmethodological,
      title={Evolutionary Multitask Optimization: a Methodological Overview, Challenges and Future Research Directions}, 
      author={Eneko Osaba and Aritz D. Martinez and Javier {Del Ser}},
      year={2021},
      eprint={2102.02558},
      archivePrefix={arXiv},
      primaryClass={cs.NE},
      url={https://arxiv.org/abs/2102.02558}
}

@article{rendell2010socialLearning,
  title   = {Why Copy Others? {{Insights}} from the Social Learning Strategies Tournament},
  author  = {Rendell, L. and Boyd, R. and Cownden, D. and Enquist, M. and Eriksson, K. and Feldman, M. W. and Fogarty, L. and Ghirlanda, S. and Lillicrap, T. and Laland, K. N.},
  journal = {Science},
  volume  = {328},
  number  = {5975},
  pages   = {208--213},
  year    = {2010},
  doi     = {10.1126/science.1184719}
}

@article{bali2020mfea2,
  title   = {Multifactorial Evolutionary Algorithm With Online Transfer Parameter Estimation: {MFEA-II}},
  author  = {Bali, Kavitesh Kumar and Ong, Yew-Soon and Gupta, Abhishek and Tan, Puay Siew},
  journal = {IEEE Transactions on Evolutionary Computation},
  volume  = {24},
  number  = {1},
  pages   = {69--83},
  year    = {2020},
  doi     = {10.1109/TEVC.2019.2906927}
}

@inproceedings{lundberg2017unified,
  title     = {A Unified Approach to Interpreting Model Predictions},
  author    = {Lundberg, Scott M. and Lee, Su-In},
  booktitle = {Advances in Neural Information Processing Systems},
  volume    = {30},
  year      = {2017}
}

@article{lundberg2020local2global,
  title   = {From Local Explanations to Global Understanding with Explainable {AI} for Trees},
  author  = {Lundberg, Scott M. and Erion, Gabriel and Chen, Hugh and DeGrave, Alex and Prutkin, Jordan M. and Nair, Bala and Katz, Ronit and Himmelfarb, Jonathan and Bansal, Nisha and Lee, Su-In},
  journal = {Nature Machine Intelligence},
  volume  = {2},
  number  = {1},
  pages   = {56--67},
  year    = {2020},
  doi     = {10.1038/s42256-019-0138-9}
}

@inproceedings{hutter2014efficient,
  title     = {An Efficient Approach for Assessing Hyperparameter Importance},
  author    = {Hutter, Frank and Hoos, Holger and Leyton-Brown, Kevin},
  booktitle = {Proceedings of the 31st International Conference on Machine Learning},
  series    = {Proceedings of Machine Learning Research},
  volume    = {32},
  pages     = {754--762},
  year      = {2014},
  publisher = {PMLR}
}

@article{mann1947test,
  title   = {On a Test of Whether one of Two Random Variables is Stochastically Larger than the Other},
  author  = {Mann, Henry B. and Whitney, Donald R.},
  journal = {The Annals of Mathematical Statistics},
  volume  = {18},
  number  = {1},
  pages   = {50--60},
  year    = {1947},
  doi     = {10.1214/aoms/1177730491}
}

@article{dunn1961multiple,
  title   = {Multiple Comparisons Among Means},
  author  = {Dunn, Olive Jean},
  journal = {Journal of the American Statistical Association},
  volume  = {56},
  number  = {293},
  pages   = {52--64},
  year    = {1961},
  doi     = {10.1080/01621459.1961.10482090}
}

@article{holm1979simple,
  title   = {A Simple Sequentially Rejective Multiple Test Procedure},
  author  = {Holm, Sture},
  journal = {Scandinavian Journal of Statistics},
  volume  = {6},
  number  = {2},
  pages   = {65--70},
  year    = {1979}
}

@book{alba2008cellular,
  title     = {Cellular Genetic Algorithms},
  author    = {Alba, Enrique and Dorronsoro, Bernab\'{e}},
  year      = {2008},
  series    = {Operations Research/Computer Science Interfaces Series},
  volume    = {42},
  publisher = {Springer US},
  address   = {Boston, MA},
  doi       = {10.1007/978-0-387-77610-1}
}

@article{bongard2006resilient,
  title   = {Resilient Machines Through Continuous Self-Modeling},
  author  = {Bongard, Josh and Zykov, Victor and Lipson, Hod},
  journal = {Science},
  volume  = {314},
  number  = {5802},
  pages   = {1118--1121},
  year    = {2006},
  doi     = {10.1126/science.1133687}
}

@article{yim2007modular,
  title   = {Modular Self-Reconfigurable Robot Systems},
  author  = {Yim, Mark and Shen, Wei-Min and Salemi, Behnam and Rus, Daniela and Moll, Mark and Lipson, Hod and Klavins, Eric and Chirikjian, Gregory S.},
  journal = {IEEE Robotics \& Automation Magazine},
  volume  = {14},
  number  = {1},
  pages   = {43--52},
  year    = {2007},
  doi     = {10.1109/MRA.2007.339623}
}

@incollection{chatzilygeroudis2021qd,
  title     = {Quality-Diversity Optimization: A Novel Branch of Stochastic Optimization},
  author    = {Chatzilygeroudis, Konstantinos and Cully, Antoine and Vassiliades, Vassilis and Mouret, Jean-Baptiste},
  booktitle = {Black Box Optimization, Machine Learning, and No-Free Lunch Theorems},
  series    = {Springer Optimization and Its Applications},
  pages     = {109--135},
  year      = {2021},
  publisher = {Springer},
  doi       = {10.1007/978-3-030-66515-9_4}
}

@article{vandiggelen2024modelfree,
  title   = {A Model-Free Method to Learn Multiple Skills in Parallel on Modular Robots},
  author  = {van Diggelen, Fuda and Cambier, Nicolas and Ferrante, Eliseo and Eiben, A. E.},
  journal = {Nature Communications},
  volume  = {15},
  pages   = {6267},
  year    = {2024},
  doi     = {10.1038/s41467-024-50131-4}
}

@article{samarakoon2022metaheuristic,
  title   = {Metaheuristic Based Navigation of a Reconfigurable Robot through Narrow Spaces with Shape Changing Ability},
  author  = {Samarakoon, S. M. Bhagya P. and Muthugala, M. A. Viraj J. and Elara, Mohan Rajesh},
  journal = {Expert Systems with Applications},
  volume  = {201},
  pages   = {117060},
  year    = {2022},
  doi     = {10.1016/j.eswa.2022.117060}
}

@misc{bartzbeielstein2026optimizationspotoptim,
      title={Optimization with SpotOptim},
      author={Thomas Bartz-Beielstein},
      year={2026},
      eprint={2604.13672},
      archivePrefix={arXiv},
      primaryClass={cs.LG},
      url={https://arxiv.org/abs/2604.13672},
}

@misc{monet_code,
  title        = {Repository for {MONET}: {M}ulti-{T}ask {O}ptimization over {N}etworks of {T}asks},
  author       = {Hatzky, Julian and Bartz-Beielstein, Thomas and Eiben, A. E. and Yaman, Anil},
  year         = {2026},
  howpublished = {\url{https://github.com/ju2ez/MONET/tree/PPSN-MONET-introduction}}
}

@misc{pymap_elites_code,
  title        = {{pymap\_elites}: {P}ython Reference Implementation of {MAP}-{E}lites and {M}ulti-{T}ask {MAP}-{E}lites},
  author       = {Mouret, Jean-Baptiste},
  howpublished = {\url{https://github.com/resibots/pymap_elites}}
}

@misc{pt_me_code,
  title        = {{P}arametric-{T}ask {MAP}-{E}lites},
  author       = {Anne, Timoth{\'e}e and Mouret, Jean-Baptiste},
  year         = {2024},
  howpublished = {\url{https://github.com/hucebot/Parametric-Task_MAP-Elites}}
}

@article{bentley1975multidimensional,
  author  = {Bentley, Jon Louis},
  title   = {Multidimensional binary search trees used for associative searching},
  journal = {Communications of the {ACM}},
  volume  = {18},
  number  = {9},
  pages   = {509--517},
  year    = {1975},
  doi     = {10.1145/361002.361007}
}
\section{Node Coverage Guarantees Under Random Sampling}

If we draw at each evaluation step a random node and perform an action, we may ask how many evaluations are required so that we visit each of the \(\nu\) nodes at least once with high certainty.

This problem can be framed as a \emph{coupon collector's problem}~\cite{feller1968introduction}. Writing $M$ for the number of draws, the expected number of steps needed to visit all \(\nu\) nodes at least once is

\[
\mathbb{E}[M]
= \nu H_\nu
= \nu \left( \ln \nu + \gamma + O\!\left(\frac{1}{\nu}\right) \right),
\]

where \(H_\nu\) is the \(\nu\)-th harmonic number and \(\gamma \approx 0.57721\) is the Euler-Mascheroni constant.

\bigskip

Beyond the expectation, one may also ask for a high-probability guarantee.
Using the asymptotic limit for the coupon collector's completion time,
the probability that all nodes have been visited by draw

\[
M = \nu(\ln \nu + c)
\]
converges to
\[
\mathbb{P}\bigl(M \le \nu(\ln \nu + c)\bigr) \approx e^{-e^{-c}}.
\]

To achieve at least \(99\%\) probability of having visited all nodes, we solve for $c$
\[
\begin{aligned}
e^{-e^{-c}} &= p \\[8pt]
\ln\!\left(e^{-e^{-c}}\right) &= \ln(p) \\[8pt]
-e^{-c} &= \ln(p) \\[8pt]
\text{For } p = 0.99:\quad
c &= -\ln\!\bigl(-\ln(0.99)\bigr) \approx 4.60015 \approx \ln(100).
\end{aligned}
\]

Since each node is selected independently with probability $\tfrac{1}{\nu}$ at every draw,
the visit count of any particular node over $M$ draws follows a Binomial distribution,
\[
X_i \sim \mathrm{Binomial}\!\left(M, \tfrac{1}{\nu}\right),
\]
from which the expected number of visits per node is
\[
\mathbb{E}[X_i] = \frac{M}{\nu}.
\]

We evaluate four environments with differing numbers of distinct tasks and a fixed amount of draws (number of evaluations) $M=1{,}000{,}000$:

\begin{table}[h]
\centering
\caption{Coupon collector bounds for the four benchmark environments with $M=1{,}000{,}000$ total evaluations.}
\label{tab:coupon-collector}
\begin{tabular}{@{}lrrrr@{}}
\toprule
Environment & $\nu$ & $\mathbb{E}[M]$ & $M_{0.99}$ & $\mathbb{E}[X_i]$ \\
\midrule
Archery / Arm / Cartpole & 5\,000 & 45\,472 & 65\,612 & 200 \\
Hexapod              & 2\,000 & 16\,356 & 24\,412 & 500 \\
\bottomrule
\end{tabular}
\end{table}

\section{MONET Algorithm}
Algorithm~\ref{alg:individual} and Algorithm~\ref{alg:social} detail the two learning subroutines invoked in Algorithm~\ref{alg:monet}.

\paragraph{Individual Learning.}
When individual learning is selected, the current elite at node $\boldsymbol{\tau}_i$ is mutated independently of any neighbor. If no solution exists at the node, a random solution is drawn from $\mathcal{U}(p_{\min}, p_{\max})$ and evaluated. Otherwise, Gaussian mutation with standard deviation $\sigma_m$ scaled by the parameter range is applied. The offspring replaces the current elite only if its fitness is at least as high.

\begin{algorithm}[H]
\caption{Individual Learning}\label{alg:individual}
\begin{algorithmic}[1]
\Require Task $\boldsymbol{\tau}$, archive $\mathcal{A}$, bounds $p_{\min}, p_{\max}$, mutation std $\sigma_m$
\Ensure Updated $\mathcal{A}[\boldsymbol{\tau}]$
\If{$\mathcal{A}[\boldsymbol{\tau}] = \emptyset$} \Comment{initialize empty node}
    \State $\boldsymbol{\theta} \sim \mathcal{U}(p_{\min}, p_{\max})$ \Comment{random solution}
    \State $\mathcal{A}[\boldsymbol{\tau}] \gets (\boldsymbol{\theta},\, f(\boldsymbol{\theta}, \boldsymbol{\tau}))$
    \State \Return
\EndIf
\State $\boldsymbol{\theta} \gets \mathcal{A}[\boldsymbol{\tau}].\text{solution}$ \Comment{current elite}
\State $\boldsymbol{\epsilon} \sim \mathcal{N}(\mathbf{0},\, \sigma_m (p_{\max} - p_{\min}))$ \Comment{Gaussian perturbation}
\State $\mathbf{y} \gets \text{clip}(\boldsymbol{\theta} + \boldsymbol{\epsilon},\; p_{\min},\; p_{\max})$ \Comment{mutant offspring}
\State $f_y \gets f(\mathbf{y}, \boldsymbol{\tau})$
\If{$f_y \geq \mathcal{A}[\boldsymbol{\tau}].\text{fitness}$} \Comment{elitist replacement}
    \State $\mathcal{A}[\boldsymbol{\tau}] \gets (\mathbf{y}, f_y)$
\EndIf
\end{algorithmic}
\end{algorithm}

\paragraph{Social Learning.}
When social learning is selected, the current elite is recombined with a solution drawn from a neighboring node. If the current node is empty, the neighbor's solution is copied as an initial seed (or a random solution is used when no neighbor has a solution). Recombination is performed via Simulated Binary Crossover (SBX)~\cite{SBX} with distribution index $\eta_c=10$~\cite{deb2001multi}. As with individual learning, the offspring replaces the elite only upon improvement.

\begin{algorithm}[H]
\caption{Social Learning}\label{alg:social}
\begin{algorithmic}[1]
\Require Task $\boldsymbol{\tau}$, neighbor task $\boldsymbol{\tau}_j$, archive $\mathcal{A}$, bounds $p_{\min}, p_{\max}$
\Ensure Updated $\mathcal{A}[\boldsymbol{\tau}]$
\If{$\mathcal{A}[\boldsymbol{\tau}] = \emptyset$} \Comment{initialize empty node}
    \If{$\mathcal{A}[\boldsymbol{\tau}_j] \neq \emptyset$}
        \State $\boldsymbol{\theta} \gets \mathcal{A}[\boldsymbol{\tau}_j].\text{solution}$ \Comment{copy from neighbor}
    \Else
        \State $\boldsymbol{\theta} \sim \mathcal{U}(p_{\min}, p_{\max})$ \Comment{random fallback}
    \EndIf
    \State $\mathcal{A}[\boldsymbol{\tau}] \gets (\boldsymbol{\theta},\, f(\boldsymbol{\theta}, \boldsymbol{\tau}))$
    \State \Return
\EndIf
\If{$\mathcal{A}[\boldsymbol{\tau}_j] = \emptyset$}
    \State \Return \Comment{no valid neighbor solution}
\EndIf
\State $\mathbf{y} \gets \text{SBX}\bigl(\mathcal{A}[\boldsymbol{\tau}].\text{solution},\; \mathcal{A}[\boldsymbol{\tau}_j].\text{solution},\; \eta_c\!=\!10\bigr)$ \Comment{crossover}
\State $\mathbf{y} \gets \text{clip}(\mathbf{y},\, p_{\min},\, p_{\max})$
\State $f_y \gets f(\mathbf{y}, \boldsymbol{\tau})$
\If{$f_y \geq \mathcal{A}[\boldsymbol{\tau}].\text{fitness}$} \Comment{elitist replacement}
    \State $\mathcal{A}[\boldsymbol{\tau}] \gets (\mathbf{y}, f_y)$
\EndIf
\end{algorithmic}
\end{algorithm}

\section{Closed-Form Task Fitness Definitions}\label{app:tasks}

This appendix gives the explicit fitness expressions for the four benchmarks.

\subsection{Archery}
With initial velocity vector
\[
\mathbf{v} = v_0
\begin{bmatrix}
-\sin(\text{yaw}) \\
\cos(\text{yaw})\cos(\text{pitch}) \\
\cos(\text{yaw})\sin(\text{pitch})
\end{bmatrix},
\quad v_0 = 70\,\mathrm{m/s},
\]
the target-plane hit time is $t_{\text{hit}} = D/v_y$ whenever $v_y > 0$; if $v_y \le 0$ the shot misses and $f = 0$. Taking $g = 9.8\,\mathrm{m/s^2}$ and wind $W\in[-10,10]\,\mathrm{m/s^2}$ along $x$, the miss coordinates on the target plane ($y = D$) are
\[
m_x = \tfrac{1}{2}\,W\,t_{\text{hit}}^2 + v_x\,t_{\text{hit}},
\qquad
m_z = -\tfrac{1}{2}\,g\,t_{\text{hit}}^2 + v_z\,t_{\text{hit}},
\qquad
r = \sqrt{m_x^2 + m_z^2}.
\]
Fitness is computed on a 10-ring target of ring width $0.061\,\mathrm{m}$:
\[
f = \frac{\max\!\bigl(0,\; 10 - \lfloor r/0.061 \rfloor\bigr)}{10} \in \{0, 0.1, \ldots, 1.0\}.
\]

\subsection{Arm}
With end-effector position $\mathbf{p}_{\text{ee}}\in\mathbb{R}^2$ and a fixed target $\mathbf{p}_{\text{tar}} = (0.5, 0.5)$,
\[
f = \exp\!\left(-\lVert \mathbf{p}_{\text{ee}} - \mathbf{p}_{\text{tar}} \rVert_2\right) \in (0,1].
\]
The joint commands are $\mathbf{q} = (\boldsymbol{\theta}-\tfrac{1}{2})\cdot 2\pi\,\alpha_{\max}/10$ where $\boldsymbol{\theta}\in[0,1]^{10}$ is the solution vector and $\alpha_{\max}$ is the task-dependent maximum joint-angle fraction; each of the $10$ links has length $L/10$ with $L$ the task-dependent total arm length. $\mathbf{p}_{\text{ee}}$ is the end-effector position returned by forward kinematics.

\subsection{Cartpole}
The task uses the \texttt{CartPole-v1} dynamics from OpenAI Gym with the pole mass $m_p$ and pole length $L$ replaced by the per-task values; the default cart mass $m_c$ is retained. With pole angle $\phi$, cart position $x_c$, control force $F\in\{-10, +10\}\,\mathrm{N}$, and gravity $g=9.8\,\mathrm{m/s^2}$,
\[
\ddot{\phi} = \frac{g \sin \phi + \cos \phi \left(-F - m_p L \dot{\phi}^2 \sin \phi\right)/(m_c + m_p)}{L \left(\tfrac{4}{3} - \tfrac{m_p \cos^2 \phi}{m_c + m_p}\right)},
\qquad
\ddot{x}_c = \frac{F + m_p L \left(\dot{\phi}^2 \sin \phi - \ddot{\phi} \cos \phi\right)}{m_c + m_p}.
\]
A solution is evaluated over $n_{\text{rollouts}} = 10$ rollouts with distinct Gym seeds. Each rollout runs until the pole leaves the balanced region ($|\phi| \ge 12^\circ$ or $|x_c| \ge 2.4\,\mathrm{m}$) or until the step cap $t_{\max}=1000$ is reached. Writing $t_{\text{balanced}}^{(k)}$ for the number of surviving steps on rollout $k$, the reported fitness is the arithmetic mean
\[
f = \frac{1}{n_{\text{rollouts}}}\sum_{k=1}^{n_{\text{rollouts}}} \min\!\bigl(t_{\max},\, t_{\text{balanced}}^{(k)}\bigr) \in [0, t_{\max}].
\]

\subsection{Hexapod}
With gait parameters $\boldsymbol{\theta}\in[0,1]^{36}$ and morphology $\boldsymbol{\tau}\in\mathcal{T}$ instantiated through a per-task URDF, a PyBullet simulation is run for up to $3\,\mathrm{s}$ of wall-clock simulator time. The episode terminates early if any body-axis angle (roll, pitch, yaw) exceeds $\pi/8$ or if the lateral deviation satisfies $|y|>0.5\,\mathrm{m}$. Writing $p_x(t)$ for the forward position of the center of mass and $T\le 3\,\mathrm{s}$ for the time of termination or episode end,
\[
f(\boldsymbol{\theta},\boldsymbol{\tau}) = p_x(T),
\]
reported in metres. The controller returns $f=0$ if any entry of $\boldsymbol{\theta}$ lies outside $[0,1]$.

\section{Detailed Sensitivity Analysis}\label{app:sensitivity}

This appendix provides per-domain SHAP beeswarm plots and the SPOT-tuned hyperparameter configurations that complement the global sensitivity analysis.

\subsection{Per-Domain SHAP Analysis}

\begin{figure}[H]
    \centering
    \includegraphics[width=1\linewidth]{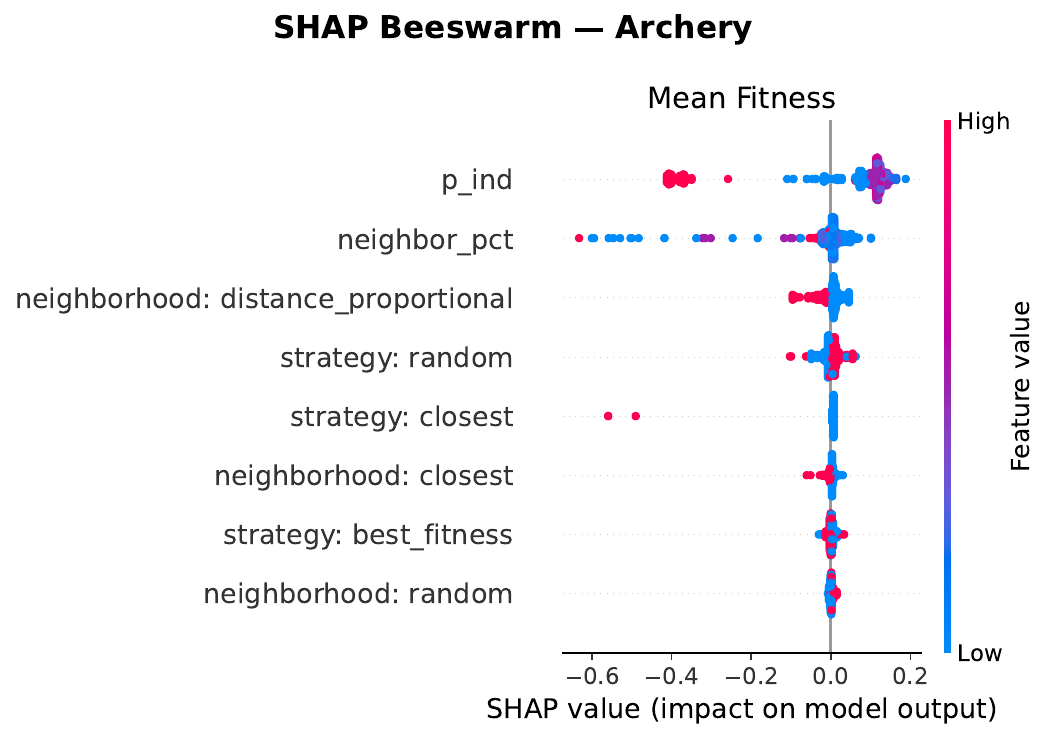}
    \caption{SHAP beeswarm plot for the archery domain. Each row is a hyperparameter; each dot is a MONET run. Color indicates the feature value (red = high, blue = low); horizontal position indicates the SHAP value.}
    \label{fig:shap-archery}
\end{figure}

\begin{figure}[H]
    \centering
    \includegraphics[width=1\linewidth]{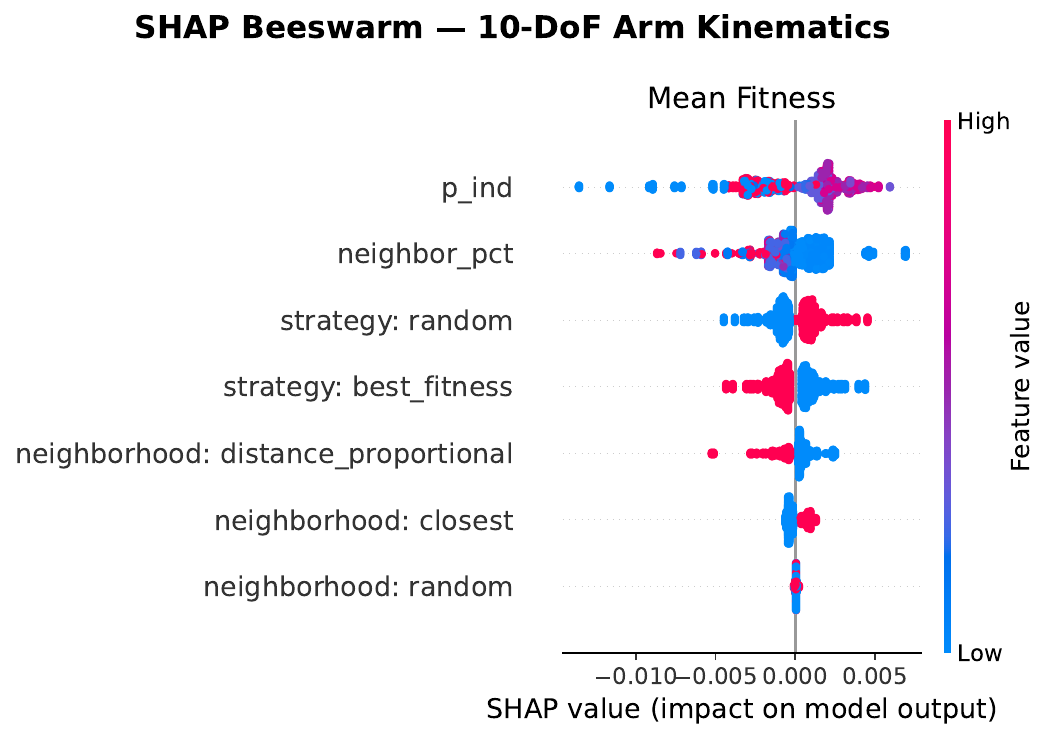}
    \caption{SHAP beeswarm plot for the arm domain. Panels as in Figure~\ref{fig:shap-archery}.}
    \label{fig:shap-arm}
\end{figure}

Figures~\ref{fig:shap-archery} and~\ref{fig:shap-arm} show the SHAP beeswarm plots for the archery and arm domains, respectively. Each point represents one MONET run; its horizontal position indicates the SHAP value (contribution to the predicted metric), and its color encodes the feature value (red = high, blue = low).

In both domains, $p_\text{ind}$ exhibits the strongest effect: high values of $p_\text{ind}$ (more individual learning) consistently decrease predicted fitness, corroborating the global finding that social learning is the primary driver of performance. The remaining parameters show smaller, more task-specific effects.

\begin{figure}[H]
    \centering
    \includegraphics[width=1\linewidth]{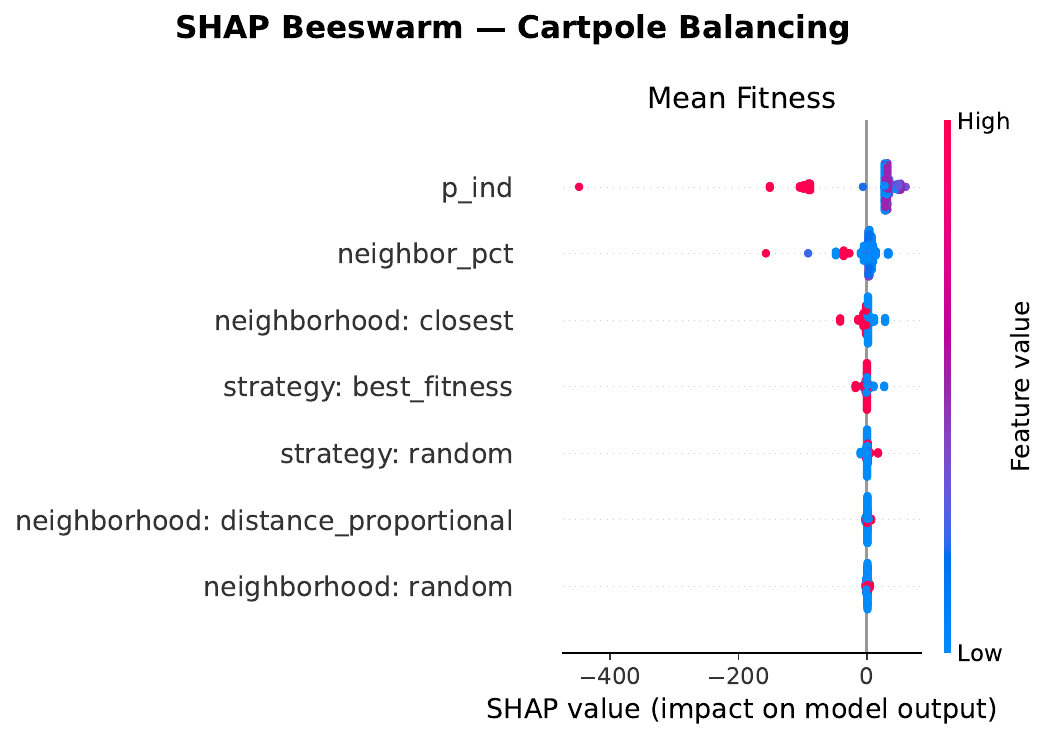}
    \caption{SHAP beeswarm plot for the cartpole domain. Panels as in Figure~\ref{fig:shap-archery}.}
    \label{fig:shap-cartpole}
\end{figure}

\begin{figure}[H]
    \centering
    \includegraphics[width=1\linewidth]{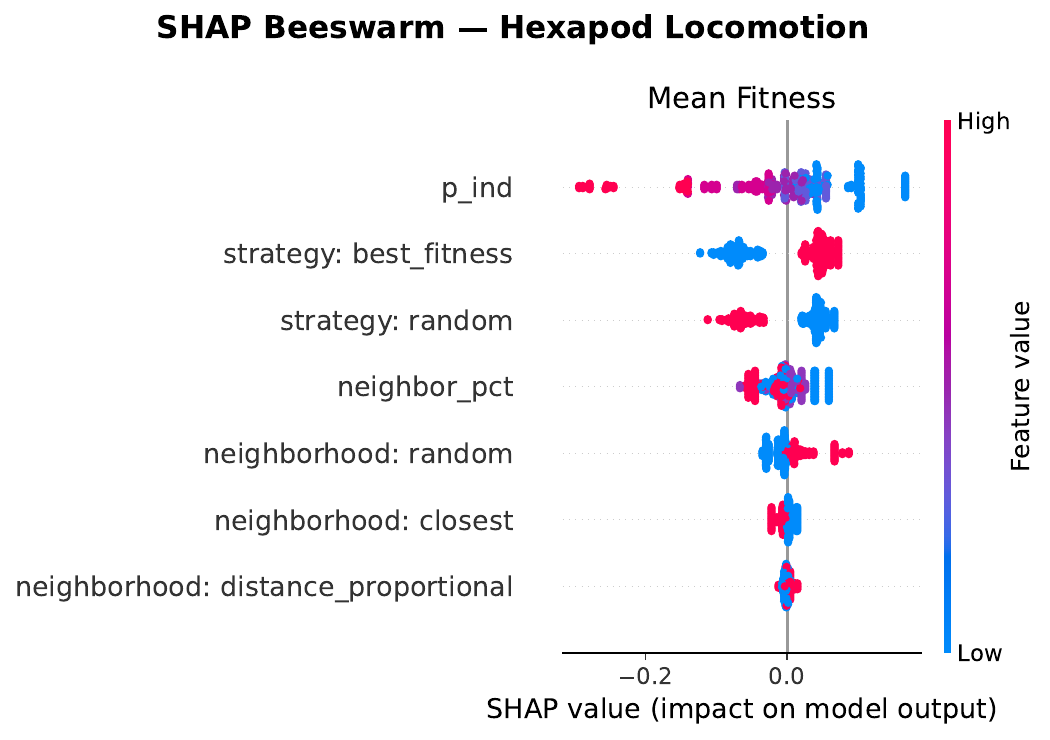}
    \caption{SHAP beeswarm plot for the hexapod domain. Panels as in Figure~\ref{fig:shap-archery}.}
    \label{fig:shap-hexapod}
\end{figure}

\subsection{SPOT-Tuned Configurations}

Table~\ref{tab:spot-configs} lists the per-domain configurations found by the SPOT hyperparameter tuner.

\begin{table}[h]
\centering
\caption{SPOT-tuned MONET configurations per domain.}
\label{tab:spot-configs}
\small
\begin{tabular}{@{}lllcc@{}}
\toprule
\textbf{Domain} & \textbf{strategy} & \textbf{neighborhood} & $p_\text{ind}$ & \textbf{neighbor\_pct.} \\
\midrule
Archery  & best\_fitness & closest  & 0.0 & 2\% \\
Arm      & random        & closest  & 0.3 & 1\% \\
Cartpole & best\_fitness & random   & 0.5 & 1\% \\
Hexapod  & best\_fitness & random   & 0.0 & 1\% \\
\bottomrule
\end{tabular}
\end{table}

\section{MONET Hyperparameter Descriptions}\label{app:hyperparameters}

Table~\ref{tab:monet-hyperparams} summarizes the four hyperparameters varied in the sensitivity analysis. All other algorithmic parameters (mutation operator, crossover operator, initialization strategy) are held fixed throughout. See Algorithm~\ref{alg:monet} for the points at which each parameter enters the loop.

\begin{table}[h]
\centering
\caption{MONET hyperparameters and their roles.}
\label{tab:monet-hyperparams}
\renewcommand{\arraystretch}{1.3}
\small
\begin{tabular}{@{}p{2.4cm}p{2.2cm}p{7.4cm}@{}}
\toprule
\textbf{Parameter} & \textbf{Values} & \textbf{Description} \\
\midrule
$p_\text{ind}$ &
0, 0.3, 0.5, 0.7, 1 &
Probability that a given evaluation step performs \emph{individual learning} (mutation of the focal task's own elite) rather than \emph{social learning} (crossover with a neighbor's elite). At $p_\text{ind}=0$ only social learning occurs; at $p_\text{ind}=1$ only individual learning occurs. \\
\addlinespace
strategy &
random, best\_fitness &
Neighbor selection strategy during social learning. \texttt{random}: uniform random selection among all neighbors. \texttt{best\_fitness}: deterministically select the neighbor with the highest current fitness. \\
\addlinespace
neighborhood &
random, closest, distance\_prop. &
Method for constructing the neighbor set of each task.
\texttt{closest}: the $N$ nearest tasks by Euclidean distance in task space.
\texttt{random}: $N$ tasks sampled uniformly at random (sorted by similarity after sampling).
\texttt{distance\_proportional}: $N$ tasks sampled \emph{without replacement} with probability proportional to task-space similarity $s(\boldsymbol{\tau}_i, \boldsymbol{\tau}_j) = 1/(1 + \lVert \boldsymbol{\tau}_i - \boldsymbol{\tau}_j \rVert)$; this softly biases the neighborhood toward nearby tasks while retaining a chance of including distant ones. \\
\addlinespace
neighbor\_pct. &
1\%, 5\%, 10\% &
Fraction of the total task set used as the neighborhood size~$N = \lceil \texttt{neighbor\_percentage} \cdot |\mathcal{V}| \rceil$. For instance, with 5\,000 tasks and 5\%, each task has $N=250$ neighbors. Controls graph sparsity: lower values yield sparser graphs with more localized knowledge transfer. \\
\bottomrule
\end{tabular}
\end{table}

\section{Hyperparameter Configuration Ranking}\label{app:ranking}

Table~\ref{tab:top10} lists the ten highest-ranked MONET hyperparameter configurations (across the 2--4 domains on which they were run) by combined z-score, computed as the mean of the within-domain z-scores of final mean fitness and AUC. Table~\ref{tab:bottom5} shows the five worst-ranked configurations for comparison. Only configurations evaluated on at least two domains are included in the ranking. The counts $n_{\text{Arm}}, n_{\text{Archery}}, n_{\text{Cartpole}}, n_{\text{Hexapod}}$ report the number of deduplicated seed runs per domain.

\begin{table}[h]
\centering
\caption{Top 10 MONET configurations by combined z-score ($n_{\text{domains}} \ge 2$).}
\label{tab:top10}
\footnotesize
\setlength{\tabcolsep}{3pt}
\begin{tabular}{@{}llrrrrrrrrr@{}}
\toprule
strategy & neighborhood & nbr.\% & $p_\text{ind}$ & $n_\text{Arm}$ & $n_\text{Arc}$ & $n_\text{Cart}$ & $n_\text{Hex}$ & $z_{\text{FF}}$ & $z_{\text{AUC}}$ & combined \\
\midrule
best\_fitness & closest               & 1.5 & 0.30 & 20 & 20 &  0 &  0 & $+0.58$ & $+1.01$ & $\mathbf{+0.80}$ \\
best\_fitness & closest               & 1.0 & 0.30 & 22 & 22 &  0 &  5 & $+0.60$ & $+0.96$ & $+0.78$ \\
best\_fitness & closest               & 2.0 & 0.30 & 20 & 20 &  0 &  0 & $+0.55$ & $+0.99$ & $+0.77$ \\
best\_fitness & closest               & 2.0 & 0.50 & 20 & 39 &  5 &  0 & $+0.62$ & $+0.90$ & $+0.76$ \\
best\_fitness & closest               & 1.5 & 0.70 &  5 & 20 &  0 &  0 & $+0.62$ & $+0.81$ & $+0.72$ \\
best\_fitness & closest               & 2.0 & 0.70 &  5 & 20 &  0 &  0 & $+0.62$ & $+0.82$ & $+0.72$ \\
random        & closest               & 2.0 & 0.00 & 20 & 19 &  4 &  0 & $+0.66$ & $+0.75$ & $+0.71$ \\
random        & closest               & 1.5 & 0.00 & 20 & 19 &  5 &  0 & $+0.66$ & $+0.75$ & $+0.70$ \\
best\_fitness & closest               & 1.0 & 0.50 & 40 & 41 &  5 &  5 & $+0.56$ & $+0.82$ & $+0.69$ \\
random        & closest               & 1.0 & 0.00 & 40 & 40 &  5 &  3 & $+0.64$ & $+0.73$ & $+0.68$ \\
\bottomrule
\end{tabular}
\end{table}

\begin{table}[h]
\centering
\caption{Bottom 5 MONET configurations by combined z-score ($n_{\text{domains}} \ge 2$).}
\label{tab:bottom5}
\footnotesize
\setlength{\tabcolsep}{3pt}
\begin{tabular}{@{}llrrrrrrrrr@{}}
\toprule
strategy & neighborhood & nbr.\% & $p_\text{ind}$ & $n_\text{Arm}$ & $n_\text{Arc}$ & $n_\text{Cart}$ & $n_\text{Hex}$ & $z_{\text{FF}}$ & $z_{\text{AUC}}$ & combined \\
\midrule
best\_fitness & distance\_prop.\       &  5.0 & 1.00 & 20 & 20 &  0 &  2 & $-1.22$ & $-1.86$ & $-1.54$ \\
best\_fitness & distance\_prop.\       & 10.0 & 0.00 & 20 & 20 &  0 & 16 & $-2.51$ & $-0.65$ & $-1.58$ \\
random        & closest               & 34.3 & 0.98 &  2 &  1 &  1 &  0 & $-1.19$ & $-1.97$ & $-1.58$ \\
best\_fitness & distance\_prop.\       &  5.0 & 0.00 & 20 & 20 &  0 &  3 & $-3.00$ & $-0.93$ & $-1.96$ \\
best\_fitness & distance\_prop.\       & 41.5 & 0.99 &  2 &  2 &  2 &  0 & $-1.97$ & $-2.23$ & $\mathbf{-2.10}$ \\
\bottomrule
\end{tabular}
\end{table}

Two patterns are clear from Table~\ref{tab:top10}: (i)~the \texttt{closest} neighborhood dominates the top of the ranking, and (ii)~small neighborhood sizes ($\le 2\%$ of the task set) are preferred. Poorly-performing configurations (Table~\ref{tab:bottom5}) combine the \texttt{distance\_proportional} neighborhood with either very high $p_\text{ind}$ (near pure individual learning) or very large neighborhoods, both of which disrupt local knowledge transfer.

\section{Code Availability}\label{app:code}

Table~\ref{tab:code-availability} lists the reference papers and public code repositories for MONET and the three baselines.

\begin{table}[h]
\centering
\caption{Reference papers and public code repositories for MONET and the baseline algorithms.}
\label{tab:code-availability}
\small
\setlength{\tabcolsep}{4pt}
\begin{tabular}{@{}lll@{}}
\toprule
\textbf{Algorithm} & \textbf{Paper} & \textbf{Code} \\
\midrule
MONET (ours) & This paper                                                    & \cite{monet_code} \\
MAP-Elites   & \cite{DBLP:journals/corr/MouretC15,MAP_Elites_NaturePaper}    & \cite{pymap_elites_code} \\
MT-ME        & \cite{mouretQualitydiversitymultitaskoptimization2020}        & \cite{pymap_elites_code} \\
PT-ME        & \cite{anneParametricTaskMAPElites2024}                        & \cite{pt_me_code} \\
\bottomrule
\end{tabular}
\end{table}

\end{document}